\documentclass[twoside]{article}
\usepackage{aistats2016}
\usepackage{times}
\usepackage{epsfig}
\usepackage{graphicx}
\usepackage{amsmath}
\usepackage{amssymb}

\usepackage{array}
\usepackage{todonotes}
\usepackage{multirow}
\usepackage{amstext}
\usepackage{graphicx}
\usepackage{subcaption}

\usepackage[breaklinks=true,bookmarks=false]{hyperref}

\usepackage{tikz}

\begin{document}

\runningtitle{Generalizing Pooling Functions in CNNs: Mixed, Gated, and Tree}

\twocolumn[

\aistatstitle{Generalizing Pooling Functions in Convolutional Neural Networks:\\ Mixed, Gated, and Tree}
\aistatsauthor{ Chen-Yu Lee \And Patrick W. Gallagher \And Zhuowen Tu }

\aistatsaddress{ UCSD ECE \And UCSD Cognitive Science \And UCSD Cognitive Science } 
\vspace{-8mm}
\aistatsaddress{ \texttt{chl260@ucsd.edu} \And \texttt{patrick.w.gallagher@gmail.com} \And \texttt{ztu@ucsd.edu}} 
]

\vspace{-7mm}
\begin{abstract}
\vspace{-3mm}
We seek to improve deep neural networks by generalizing the pooling operations 
that play a central role in current architectures. 
We pursue a careful exploration of approaches to allow pooling to 
learn and to adapt to complex and variable 
patterns. The two primary directions lie in (1) learning a pooling function via 
(two strategies of) combining of max and average pooling, and (2) learning a pooling 
function in the form of a tree-structured fusion of pooling filters that are themselves 
learned. In our experiments every generalized pooling operation we 
explore improves performance when used in place of average or max pooling.  
We experimentally demonstrate that the proposed pooling operations provide a boost 
in invariance properties relative to conventional pooling and 
set the state of the art on several widely adopted benchmark datasets; they are also easy to 
implement, and can be applied within various deep neural network architectures.
These benefits come with only a light increase in computational overhead during training 
and a very modest increase in the number of model parameters. 
\end{abstract}

\vspace{-6mm}
\section{Introduction}
\vspace{-3mm}

The recent resurgence of neurally-inspired systems such as deep belief nets (DBN) 
\cite{hinton2006fast}, convolutional neural networks (CNNs) \cite{lecun1989backpropagation},
and the sum-and-max infrastructure \cite{serre2007robust} has derived significant benefit 
from building more sophisticated network structures \cite{szegedy2014going, simonyan2015very} and 
from bringing learning to non-linear activations \cite{goodfellow2013maxout, lin2013network}. 
The pooling operation has also played a central role, contributing 
to invariance to data variation and perturbation.
However, pooling operations have been little revised beyond the current primary options of
average, max, and stochastic pooling \cite{boureau2010theoretical,zeiler2013stochastic};
this despite indications that e.g. choosing from more than just one type of 
pooling operation can benefit performance \cite{scherer2010evaluation}.

In this paper, we desire to bring learning and ``responsiveness'' (i.e., to characteristics 
of the region being pooled) into the pooling operation. 
Various approaches are possible, but here we pursue two in particular. 
In the first approach, we consider combining typical pooling operations (specifically,
max pooling and average pooling); within this approach we further investigate 
two strategies by which to combine these operations. One of the strategies is
``unresponsive''; for reasons discussed later, we call this strategy 
\emph{mixed max-average pooling}. The other strategy is ``responsive''; we call 
this  strategy \emph{gated max-average pooling}, where the ability to be responsive 
is provided by a ``gate'' in analogy to the usage of gates elsewhere in deep learning.

Another natural generalization of pooling operations is to allow the pooling operations 
that are being combined to themselves be learned. Hence in the second approach, we 
learn to combine pooling filters that are themselves learned. Specifically,
the learning is performed within a binary tree (with number of levels that is 
pre-specified rather than ``grown'' as in traditional decision trees) in which each 
leaf is associated with a learned pooling filter. As we consider internal nodes 
of the tree, each parent node is associated with an output value that is the mixture 
of the child node output values, until we finally reach the root node. The root
node corresponds to the overall output produced by the tree. We refer to this strategy as \emph{tree 
pooling}. Tree pooling is intended (1) to learn pooling filters directly from the data;
(2) to learn how to combine leaf node pooling filters in a differentiable fashion; 
(3) to bring together these other characteristics within a hierarchical tree structure.

When the mixing of the node outputs is allowed to be ``responsive'',
the resulting tree pooling operation becomes an integrated method for learning pooling 
filters and combinations of those filters that are able to display a range of different behaviors 
depending on the characteristics of the region being pooled.

We pursue experimental validation and find that: In the architectures we investigate, 
replacing standard pooling operations with any of our proposed generalized pooling methods 
boosts performance on each of the standard benchmark datasets, as well as on the larger and 
more complex ImageNet dataset. We attain state-of-the-art results
on MNIST, CIFAR10 (with and without data augmentation), and SVHN. Our proposed pooling operations 
can be used as drop-in replacements for standard pooling operations in various current architectures 
and can be used in tandem with other performance-boosting approaches such as learning activation 
functions, training with data augmentation, or modifying other aspects of network architecture --- 
we confirm improvements when used in a DSN-style architecture, as well as in AlexNet and GoogLeNet. 
Our proposed pooling operations are also simple to implement, computationally undemanding (ranging from $5\%$ 
to $15\%$ additional overhead in timing experiments), differentiable, and use 
only a modest number of additional parameters.
\vspace{-4mm}
\section{Related Work}
\vspace{-4mm}

In the current deep learning literature, popular pooling functions include max,
average, and stochastic pooling \cite{boureau2010theoretical,boureau2011ask,zeiler2013stochastic}.
A recent effort using more complex pooling operations, spatial pyramid pooling 
\cite{he2014spatial}, is mainly designed to deal with images of varying size, 
rather than delving in to different pooling functions or incorporating 
learning. Learning pooling functions is analogous to 
receptive field learning \cite{gulcehre2014learned,hubel1962receptive,coates2011selecting,
jia2012beyond}. However methods like \cite{jia2012beyond} lead to a more 
difficult learning procedure that in turn leads to a less competitive result, 
e.g. an error rate of $16.89\%$ on unaugmented CIFAR10. 

Since our tree pooling approach involves a tree structure in its learning, 
we observe an analogy to ``logic-type'' approaches such as decision trees \cite{quinlan1993c45} 
or ``logical operators'' \cite{minker2000logic}. Such approaches have played a central 
role in artificial intelligence for applications that require ``discrete'' reasoning,
and are often intuitively appealing. Unfortunately, despite the appeal of such logic-type 
approaches, there is a disconnect between the functioning of decision trees and 
the functioning of CNNs --- the output of a standard decision tree is non-continuous 
with respect to its input (and thus nondifferentiable). This means that a standard 
decision tree is not able to be used in CNNs, whose learning process is performed 
by back propagation using gradients of differentiable functions. Part of what allows 
us to pursue our approaches is that we ensure the resulting pooling operation is 
differentiable and thus usable within network backpropagation. 

A recent work, referred to as auto-encoder trees \cite{irsoy2014autoencoder}, also
pays attention to a differentiable use of tree structures in deep learning, but 
is distinct from our method as it focuses on learning encoding and decoding methods 
(rather than pooling methods) using a ``soft'' decision tree for a generative model. 
In the supervised setting, \cite{bulo2014neural} incorporates multilayer perceptrons 
within decision trees, but simply uses trained perceptrons as splitting nodes in 
a decision forest; not only does this result in training processes that are separate
(and thus more difficult to train than an integrated training process), this training 
process does not involve the learning of any pooling filters.

\vspace{-4mm}
\section{Generalizing Pooling Operations}
\vspace{-4mm}
A typical convolutional neural network is structured as a series of convolutional
layers and pooling layers. Each convolutional layer is intended to produce representations
(in the form of activation values) that reflect aspects of local spatial structures, 
and to consider multiple channels  when doing so. More specifically, a convolution layer computes ``feature response maps''
that involve multiple channels within some localized spatial region. On the other hand, 
a pooling layer is restricted to act within just one channel at a time, ``condensing'' the 
activation values in each spatially-local region in the currently considered channel. An 
early reference related to pooling operations (although not explicitly using the term ``pooling'')
can be found in \cite{hubel1962receptive}. In modern visual recognition systems, pooling operations 
play a role in producing ``downstream'' representations that are more robust to the effects of 
variations in data while still preserving important motifs. The specific choices of average pooling \cite{lecun1989backpropagation,lecun1998gradientbased} 
and max pooling \cite{ranzato2007sparse} have been widely used in many CNN-like
architectures; \cite{boureau2010theoretical} includes a theoretical analysis (albeit one
based on assumptions that do not hold here).

Our goal is to bring learning and ``responsiveness'' into the pooling operation.
We focus on two approaches in particular. In the first approach, we begin with 
the (conventional, non-learned) pooling operations of max pooling and average pooling and learn to combine 
them. Within this approach, we further consider two strategies by which to combine these
fixed pooling operations. One of these strategies is ``unresponsive'' to the characteristics
of the region being pooled; the learning process in this strategy will result in
an effective pooling operation that is some specific, unchanging ``mixture'' of max 
and average. To emphasize this unchanging mixture, we refer to this strategy as 
\emph{mixed max-average pooling}. 

The other strategy is ``responsive'' to the characteristics of the region being pooled; 
the learning process in this strategy results in a ``gating mask''. This learned 
gating mask is then used to determine a ``responsive'' mix of max pooling and average 
pooling; specifically, the value of the inner product between the gating mask and 
the current region being pooled is fed through a sigmoid, the output of which is 
used as the mixing proportion between max and average. To emphasize the role of 
the gating mask in determining the ``responsive'' mixing proportion, we refer to 
this strategy as \emph{gated max-average pooling}. 

Both the mixed strategy and the gated strategy involve combinations of fixed pooling
operations; a complementary generalization to these strategies is to learn 
the pooling operations themselves. From this, we are in turn led to consider learning pooling 
operations and also learning to combine those pooling operations. Since these combinations 
can be considered within the context of a binary tree structure, we refer to 
this approach as \emph{tree pooling}. We pursue further details in the following 
sections.

\vspace{-4mm}
\subsection{Combining max and average pooling functions}
\vspace{-2.7mm}

\subsubsection{``Mixed'' max-average pooling}
\vspace{-2.7mm}
The conventional pooling operation is fixed to be either a simple average
$ f_{\text{ave}}({\mathbf{x}})=\frac{1}{N}\sum^{N}_{i=1}{{\mathbf{x}}_i} $ or
a maximum operation $f_{\text{max}}({\mathbf{x}})=\max_i{{\mathbf{x}}_i}$, 
where the vector ${\mathbf{x}}$ contains the  
activation values from a local pooling region of $N$ pixels (typical pooling 
region dimensions are $2\times2$ or $3\times3$) in an image or a channel. 

At present, max pooling is often used as the default in CNNs. We touch on the relative 
performance of max pooling and, e.g., average pooling as part 
of a collection of exploratory experiments to test the invariance properties of pooling
functions under common image transformations (including rotation, translation, 
and scaling); see Figure \ref{fig:invariance}. The results indicate that, on the 
evaluation dataset, there are regimes in which either max pooling or average pooling 
demonstrates better performance than the other (although we observe that both of 
these choices are outperformed by our proposed pooling operations). In the light 
of observation that neither max pooling nor average pooling dominates the other, 
a first natural generalization is the strategy we call ``mixed'' max-average pooling,
in which we learn specific mixing proportion parameters from the data. When 
learning such mixing proportion parameters one has several options (listed in order of
increasing number of parameters): learning one mixing proportion parameter (a) per 
net, (b) per layer, (c) per layer/region being pooled (but used for all channels across 
that region), (d) per layer/channel (but used for all regions in each channel) 
(e) per layer/region/channel combination.

\begin{figure}[t]
  \centering
  \includegraphics[width=0.9\linewidth]{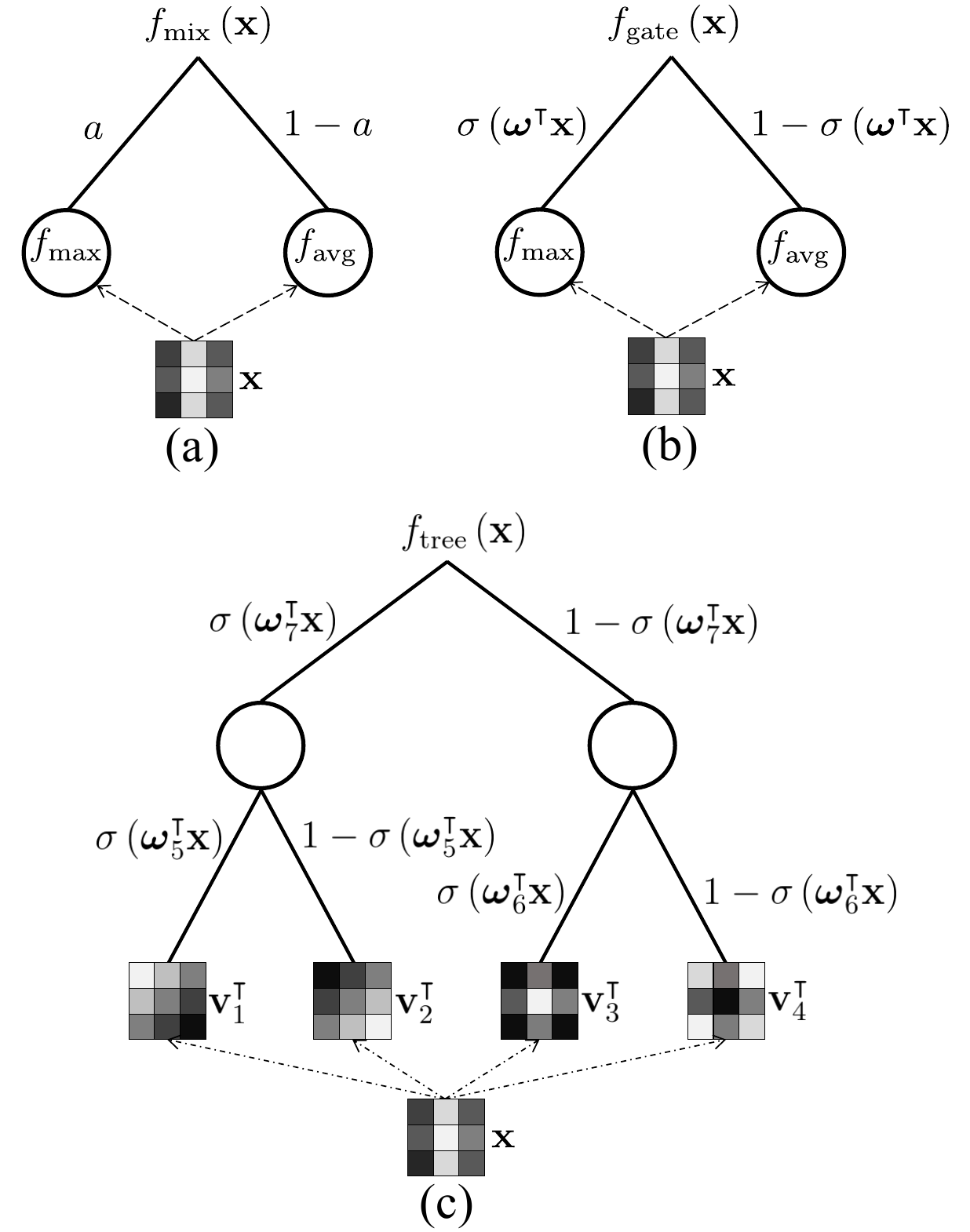}
\vspace{-2mm}
\caption{Illustration of proposed pooling operations: (a) mixed max-average pooling, (b) gated max-average pooling, and (c) Tree pooling (3 levels in this figure).  We indicate the region being pooled by $\mathbf{x}$, gating masks by $\boldsymbol{\omega}$, and pooling filters by $\mathbf{v}$ (subscripted as appropriate). }
\label{fig:pool}
\vspace{-3.4mm}
\end{figure}

The form for each ``mixed'' pooling operation (written here for the ``one per layer'' 
option; the expression for other options differs only in the subscript of the mixing 
proportion $a$) is:
\vspace{-3mm}
{\small
\begin{equation}
   f_{\text{mix}}(\mathbf{\mathbf{x}}) = a_{\ell} \cdot f_{\text{max}}({\mathbf{x}})
   + (1-a_{\ell}) \cdot f_{\text{avg}}({\mathbf{x}})
   \vspace{-2.1mm}
\end{equation}
}%
where $a_{\ell} \in [0,1]$ is a scalar mixing proportion specifying the specific 
combination of max and average; the subscript $\ell$ is used to indicate that this equation 
is for the ``one per layer'' option.
Once the output loss function $E$ is defined, we can automatically learn each 
mixing proportion $a$ (where we now suppress any subscript specifying which of 
the options we choose). Vanilla backpropagation for this learning is given by 
\vspace{-2mm}
{\small
\begin{equation}
      \frac{\partial E}{\partial a}
    = \frac{\partial E}{\partial f_{\text{mix}}(\mathbf{x})} 
         \frac{\partial f_{\text{mix}}(\mathbf{x})}{\partial a} 
      = \delta \: (\max_i{{\mathbf{x}}_i} - \frac{1}{N}\sum^{N}_{i=1}{{\mathbf{x}}_i}),
\end{equation}
}%
where $\delta = \partial E / \partial f_{\text{mix}}(\mathbf{x})$ is the error backpropagated
from the following layer. Since pooling operations are typically placed in the midst
of a deep neural network, we also need to compute the error signal to be propagated
back to the previous layer:
\vspace{-2mm}
{\small
\begin{align}
      \frac{\partial E}{\partial \mathbf{x}_{i}}
    &= \frac{\partial E}{\partial f_\text{mix}(\mathbf{x}_{i})} 
         \frac{\partial f_\text{mix}(\mathbf{x}_{i})}{\partial \mathbf{x}_{i}} \\    
      &= \delta \: \left[a \cdot \mathbf{1}[\mathbf{x}_{i} = \max_i{{\mathbf{x}}_i}] + (1-a)\cdot \frac{1}{N}\right],
      \vspace{-2.5mm}
\end{align}
}%
where $\mathbf{1}[\cdot]$ denotes the $0/1$ indicator function. In the experiment 
section, we report results for the ``one parameter per pooling layer'' option; the  
network for this experiment has 2 pooling layers and so has 2 more parameters than 
a network using standard pooling operations. We found that even this simple option 
yielded a surprisingly large performance boost. We also obtain results for a simple 50/50 mix of max and average, 
as well as for the option with the largest number of parameters: one parameter for each 
combination of layer/channel/region, or $pc \times ph \times pw$ parameters for each ``mixed'' pooling layer using this
option (where $pc$ is the number of channels being pooled by the pooling layer, 
and the number of spatial regions being pooled in each channel is $ph\times pw$). 
We observe that the increase in the number of parameters is not met with 
a corresponding boost in performance, and so we pursue the ``one per layer'' option.

\subsubsection{``Gated'' max-average pooling}
\vspace{-2mm}
In the previous section we considered a strategy that we referred to as 
``mixed'' max-average pooling; in that strategy we learned a mixing proportion to 
be used in combining max pooling and average pooling. As mentioned earlier, once
learned, each mixing proportion $a$ remains fixed --- it is ``nonresponsive'' insofar 
as it remains the same no matter what characteristics are present in the region 
being pooled. We now consider a ``responsive'' strategy that we call ``gated'' max-average
pooling. In this strategy, rather than directly learning a mixing proportion that will be 
fixed after learning, we instead learn a ``gating mask'' (with spatial dimensions matching 
that of the regions being pooled). The scalar result of the inner product between 
the gating mask and the region being pooled is fed through a sigmoid to produce 
the value that we use as the mixing proportion.  This strategy means that the actual 
mixing proportion can vary during use depending on characteristics present in the 
region being pooled. To be more specific, suppose we use $\mathbf{x}$ to denote 
the values in the region being pooled and $\boldsymbol{\omega}$ to denote the values 
in a ``gating mask''. The ``responsive'' mixing proportion is then given by
$\sigma (\boldsymbol{\omega}^\intercal \mathbf{x})$, where
$\sigma(\boldsymbol{\omega}^\intercal \mathbf{x}) = 1/(1+\exp\{ -\boldsymbol{\omega}^\intercal \mathbf{x} \} ) \in [0, 1]$ 
is a sigmoid function. 

Analogously to the strategy of learning mixing proportion parameter, when learning 
gating masks one has several options (listed in order of increasing number of parameters):
learning one gating mask (a) per net, (b) per layer, (c) per layer/region being pooled 
(but used for all channels across that region), (d) per layer/channel (but used for all 
regions in each channel) (e) per layer/region/channel combination. We suppress the 
subscript denoting the specific option, since the equations are otherwise identical for each option.

The resulting pooling operation for this ``gated'' max-average pooling is:  
{\footnotesize
\begin{equation}
   f_{\text{gate}}(\mathbf{\mathbf{x}}) = \sigma (\boldsymbol{\omega}^\intercal \mathbf{x}) f_{\text{max}}({\mathbf{x}})   + (1-\sigma(\boldsymbol{\omega}^\intercal \mathbf{x})) f_{\text{avg}}({\mathbf{x}})
\end{equation}
}%
We can compute the gradient with respect to the internal ``gating mask'' $\boldsymbol{\omega}$ 
using the same procedure considered previously, yielding
{\scriptsize
\begin{align}
      \frac{\partial E}{\partial \boldsymbol{\omega}}
     &= \frac{\partial E}{\partial f_{\text{gate}}(\mathbf{x})} 
         \frac{\partial f_{\text{gate}}(\mathbf{x})}{\partial \boldsymbol{\omega}} \\
       &= \delta \: \sigma(\boldsymbol{\omega}^\intercal \mathbf{x}) (1-\sigma(\boldsymbol{\omega}^\intercal \mathbf{x})) \: \mathbf{x} \: (\max_i{{\mathbf{x}}_i} - \frac{1}{N}\sum^{N}_{i=1}{{\mathbf{x}}_i}),
\end{align}
}%
and
{\scriptsize
\begin{align}
\frac{\partial E}{\partial \mathbf{x}_{i}}  &= \frac{\partial E}{\partial f_\text{gate}(\mathbf{x}_{i})} 
         \frac{\partial f_\text{gate}(\mathbf{x}_{i})}{\partial \mathbf{x}_{i}} \\
&= \delta \: \bigg[\sigma(\boldsymbol{\omega}^\intercal \mathbf{x}) (1-\sigma(\boldsymbol{\omega}^\intercal \mathbf{x})) \: \boldsymbol{\omega}_{i} \: (\max_i{{\mathbf{x}}_i} - \frac{1}{N}\sum^{N}_{i=1}{{\mathbf{x}}_i}) \\
& \hspace{2.7mm}  + \sigma(\boldsymbol{\omega}^\intercal \mathbf{x})\cdot\mathbf{1}[\mathbf{x}_{i} = \max_i{{\mathbf{x}}_i}] + (1-\sigma(\boldsymbol{\omega}^\intercal \mathbf{x})) \frac{1}{N}\bigg]. \nonumber
\end{align}
}%
In a head-to-head parameter count, every single mixing proportion parameter $a$ in the ``mixed''
max-average pooling strategy corresponds to a gating mask $\boldsymbol{\omega}$ in the ``gated'' strategy (assuming they use 
the same parameter count option). To take a specific example, suppose that we consider a 
network with 2 pooling layers and pooling regions that are $3 \times 3$. If we use the ``mixed'' 
strategy and the per-layer option, we would have a total of $2 = 2\times 1$ extra parameters relative
to standard pooling. If we use the ``gated'' strategy and the per-layer option, 
we would have a total of $18 = 2\times 9$ extra parameters, where $9$ is the number 
of parameters in each gating mask. The ``mixed'' strategy detailed immediately above uses fewer parameters and  
is ``nonresponsive''; the ``gated'' strategy involves more parameters and is ``responsive''. 
In our experiments, we find that ``mixed'' (with one mix per pooling layer) is outperformed
by ``gated'' with one gate per pooling layer. Interestingly, an $18$ parameter ``gated'' network with only one gate 
per pooling layer also outperforms  a ``mixed'' option with far more parameters ($40,\!960$ with one mix per 
layer/channel/region) --- except on the relatively 
large SVHN dataset. We touch on this below; Section \ref{sec:exp} contains details.
\vspace{-3mm}
\subsubsection{Quick comparison: mixed and gated pooling}
\vspace{-2mm}
The results in Table \ref{tab:mixed_and_gated} indicate the benefit of learning pooling operations
over not learning. Within learned pooling operations, we see that when the number of 
parameters in the mixed strategy is increased, performance improves; however, parameter 
count is not the entire story. We see that the ``responsive'' gated max-avg strategy 
consistently yields better performance (using 18 extra parameters) than is achieved with the $>$40k
extra parameters in the 1 per layer/rg/ch ``non-responsive'' mixed max-avg strategy. The relatively 
larger SVHN dataset provides the sole exception (SVHN has $\approx$600k training images versus 
$\approx$50k for MNIST, CIFAR10, and CIFAR100) --- we found baseline 1.91\%, 50/50 mix 1.84\%, mixed (1 per lyr) 1.76\%, mixed (1 per lyr/ch/rg) 1.64\%, and gated (1 per lyr) 1.74\%.

\begin{table}[!htp]
\footnotesize
\vspace{-2mm}
\caption{\label{tab:mixed_and_gated} Classification error (in \%) comparison between baseline model (trained
with conventional max pooling) and corresponding networks in which max pooling is 
replaced by the pooling operation listed. A superscripted $^{+}$ indicates the standard data augmentation 
as in  \cite{lin2013network,lee2015deeply,springenberg2015striving}. We report means and standard deviations 
over 3 separate trials without model averaging.}
\vspace{-2mm}
\begin{center}
\small
\begin{tabular}{c | c | c | c | c }
\hline
\shortstack{Method}                                      & {\tiny MNIST}   & {\tiny CIFAR10} & {\tiny CIFAR10$^{+}$}  & {\tiny CIFAR100} \\
\hline
Baseline                                                    & 0.39            & 9.10            & 7.32                   & 34.21            \\
\hline
\shortstack{w/ \textbf{Stochastic}  \\ {\tiny no learning}}  & \shortstack{ 0.38\\ {\tiny $\:\pm$ 0.04}}            & \shortstack{ 8.50\\ {\tiny $\:\pm$ 0.05}}            & \shortstack{ 7.30\\ {\tiny $\:\pm$ 0.07}}                & \shortstack{ 33.48\\ {\tiny $\:\pm$ 0.27}}            \\
\hline
\shortstack{w/ \textbf{50/50 mix} \\ {\tiny no learning}}   & \shortstack{ 0.34\\ {\tiny $\:\pm$ 0.012}}             &  \shortstack{ 8.11\\ {\tiny $\:\pm$ 0.10}}            & \shortstack{ 6.78\\ {\tiny $\:\pm$ 0.17}}  & \shortstack{ 33.53\\ {\tiny $\:\pm$ 0.16}}             \\
\hline
\shortstack{w/ \textbf{Mixed} \\ {\tiny 1 per pool layer } \\{\tiny 2 extra params} }  &  \shortstack{ 0.33\\ {\tiny $\:\pm$ 0.018}}             &  \shortstack{ 8.09 \\ {\tiny $\:\pm$ 0.19}}            &  \shortstack{ 6.62\\ {\tiny $\:\pm$ 0.21}}                   &  \shortstack{ 33.51\\ {\tiny $\:\pm$ 0.11}}             \\
\hline
\shortstack{w/ \textbf{Mixed} \\ {\tiny 1 per layer/ch/rg} \\{\tiny $>$40k extra params} } &\shortstack{ 0.30 \\ {\tiny $\:\pm$ 0.012}}  & \shortstack{ 8.05 \\ {\tiny $\:\pm$ 0.16}}    &  \shortstack{ 6.58 \\ {\tiny $\:\pm$ 0.30}}                   &  \shortstack{ 33.35 \\ {\tiny $\:\pm$ 0.19}}            \\
\hline
\shortstack{w/ \textbf{Gated} \\ {\tiny 1 per pool layer} \\ {\tiny 18 extra params} }  & \shortstack{\textbf{0.29} \\ {\tiny $\:\pm$ 0.016}}   & \shortstack{\textbf{7.90} \\ {\tiny $\:\pm$ 0.07}}   & \shortstack{\textbf{6.36} \\ {\tiny $\:\pm$ 0.28}}          & \shortstack{\textbf{33.22} \\ {\tiny $\:\pm$ 0.16}}   \\
\hline
\end{tabular}
\end{center}
\vspace{-4mm}
\end{table}
\vspace{-2mm}
\subsection{Tree pooling}
\vspace{-2mm}
The strategies described above each involve combinations of fixed pooling operations;
another natural generalization of pooling operations is to allow the pooling operations
that are being combined to themselves be learned.  These pooling layers remain distinct from convolution 
layers since pooling is performed separately within each channel; this channel isolation also means 
that even the option that introduces the largest number of parameters still introduces far fewer parameters 
than a convolution layer would introduce. The most basic version of this 
approach would not involve combining learned pooling operations, but simply learning pooling operations 
in the form of the values in “pooling filters”. One step further brings us to what 
we refer to as \emph{tree pooling}, in which we learn pooling filters and also learn 
to responsively combine those learned filters. 

Both aspects of this learning are performed within a binary tree (with number of levels that is 
pre-specified rather than ``grown'' as in traditional decision trees) in which each 
leaf is associated with a pooling filter learned during training. As we consider internal nodes 
of the tree, each parent node is associated with an output value that is the mixture 
of the child node output values, until we finally reach the root node. The root
node corresponds to the overall output produced by the tree and each of the mixtures 
(by which child outputs are ``fused'' into a parent output) is responsively learned. 
Tree pooling is intended (1) to learn pooling filters directly from the data;
(2) to learn how to ``mix'' leaf node pooling filters in a differentiable fashion; 
(3) to bring together these other characteristics within a hierarchical tree structure.

Each leaf node in our tree is associated with a ``pooling filter'' that will be learned; 
for a node with index $m$, we denote the pooling filter  by $\mathbf{v}_{m} \in \mathbb{R}^N$.
If we had a ``degenerate tree'' consisting of only a single (leaf) node, pooling a region 
$\mathbf{x} \in \mathbb{R}^N$ would result in the scalar value $ \mathbf{v}_{m}^\intercal \mathbf{x}$. 
For (internal) nodes (at which two child values are combined into a single 
parent value), we proceed in a fashion analogous to the case of gated max-average 
pooling, with learned ``gating masks'' denoted (for an internal node $m$) by 
$\boldsymbol{\omega}_{m} \in \mathbb{R}^N$. The ``pooling result'' at any arbitrary
node $m$ is thus 
\vspace{-2mm}
{\tiny
\begin{equation}
   f_{m}(\mathbf{x})=
   \begin{cases}
  \mathbf{v}^\intercal_{m} \mathbf{x} & \text{ if leaf node} \\
   \sigma (\boldsymbol{\omega}^\intercal_{m} \mathbf{x}) f_{{m},\text{left}}({\mathbf{x}}) + (1-\sigma(\boldsymbol{\omega}^\intercal_{m} \mathbf{x})) f_{{m},\text{right}}({\mathbf{x}}) & \text{ if internal node}
   \end{cases}
\end{equation}
}%
The overall pooling operation would thus be the result of evaluating $f_{\text{root\_node}}(\mathbf{x})$.
The appeal of this tree pooling approach would be limited if one could not train the 
proposed layer in a fashion that was integrated within the network as a whole. This
would be the case if we attempted to directly use a traditional decision tree, since
its output presents points of discontinuity with respect to its inputs. The reason 
for the discontinuity (with respect to input) of traditional decision tree output 
is that a decision tree makes ``hard'' decisions; in the terminology we have used above, 
a ``hard'' decision node corresponds to a mixing proportion that can only take on 
the value $0$ or $1$. The consequence is that this type of ``hard'' function is 
not differentiable (nor even continuous with respect to its inputs), and this in 
turn interferes with any ability to use it in iterative parameter updates during 
backpropagation. This motivates us to instead use the internal node sigmoid ``gate''
function $\sigma(\boldsymbol{\omega}_{m}^\intercal \mathbf{x})\in [0, 1]$ so that the tree 
pooling function as a whole will be differentiable with respect to its parameters 
and its inputs. 

For the specific case of a ``2 level'' tree (with leaf nodes ``1'' and ``2'' and internal node ``3'')  
pooling function $f_\text{tree}(\mathbf{\mathbf{x}}) = \sigma(\boldsymbol{\omega}_{3}^\intercal \mathbf{x} ) \mathbf{v}_{1}^\intercal \mathbf{x} + (1-\sigma(\boldsymbol{\omega}_{3}^\intercal \mathbf{x} )) \mathbf{v}^\intercal_{2} \mathbf{x} $, 
we can use the chain rule to compute the gradients 
with respect to the leaf node pooling filters $\mathbf{v}_{1},\mathbf{v}_{2}$ and 
the internal node gating mask $\boldsymbol{\omega}_{3}$:
{
\tiny
\begin{equation}
      \frac{\partial E}{\partial \mathbf{v}_\mathbf{1}}
    = \frac{\partial E}{\partial f_\text{tree}(\mathbf{x})} 
         \frac{\partial f_\text{tree}(\mathbf{x})}{\partial \mathbf{v}_1} 
      = \delta \: \sigma (\boldsymbol{\omega}_{3}^\intercal \mathbf{x}) \mathbf{x}
\end{equation}    
\begin{equation}
      \frac{\partial E}{\partial \mathbf{v}_2}
    = \frac{\partial E}{\partial f_\text{tree}(\mathbf{x})} 
         \frac{\partial f_\text{tree}(\mathbf{x})}{\partial \mathbf{v}_2} 
      = \delta \: (1-\sigma (\boldsymbol{\omega}_{3}^\intercal \mathbf{x})) \mathbf{x}
\end{equation} 
\begin{equation}
      \frac{\partial E}{\partial \boldsymbol{\omega}_{3}}
    = \frac{\partial E}{\partial f_\text{tree}(\mathbf{x})} 
         \frac{\partial f_\text{tree}(\mathbf{x})}{\partial \boldsymbol{\omega}_{3}} 
      = \delta \: \sigma (\boldsymbol{\omega}_{3}^\intercal \mathbf{x})(1-\sigma (\boldsymbol{\omega}_{3}^\intercal \mathbf{x})) \: \mathbf{x} \: (\mathbf{v}^\intercal_{1}  - 
        \mathbf{v}^\intercal_{2}  ) \mathbf{x}
\end{equation} 
}%
The error signal to be propagated back to the previous layer is
{\scriptsize
\begin{align}
      \frac{\partial E}{\partial \mathbf{x}}
    &= \frac{\partial E}{\partial f_\text{tree}(\mathbf{x})} 
         \frac{\partial f_\text{tree}(\mathbf{x})}{\partial \mathbf{x}} \\
      &= \delta \: [ \sigma (\boldsymbol{\omega}_{3}^\intercal \mathbf{x}) (1-\sigma (\boldsymbol{\omega}_{3}^\intercal \mathbf{x})) \boldsymbol{\omega}_{3} ( \mathbf{v}^\intercal_1 - \mathbf{v}^\intercal_2 ) \mathbf{x} \\
      &\hspace{2.7mm}  +  \sigma (\boldsymbol{\omega}_{3}^\intercal \mathbf{x})\mathbf{v}_1 + (1-\sigma (\boldsymbol{\omega}_{3}^\intercal \mathbf{x})) \mathbf{v}_2  ] \nonumber
\end{align} 
}%
\vspace{-5mm}
\begin{figure*}[t]
\begin{center}
\includegraphics[width=0.33\linewidth]{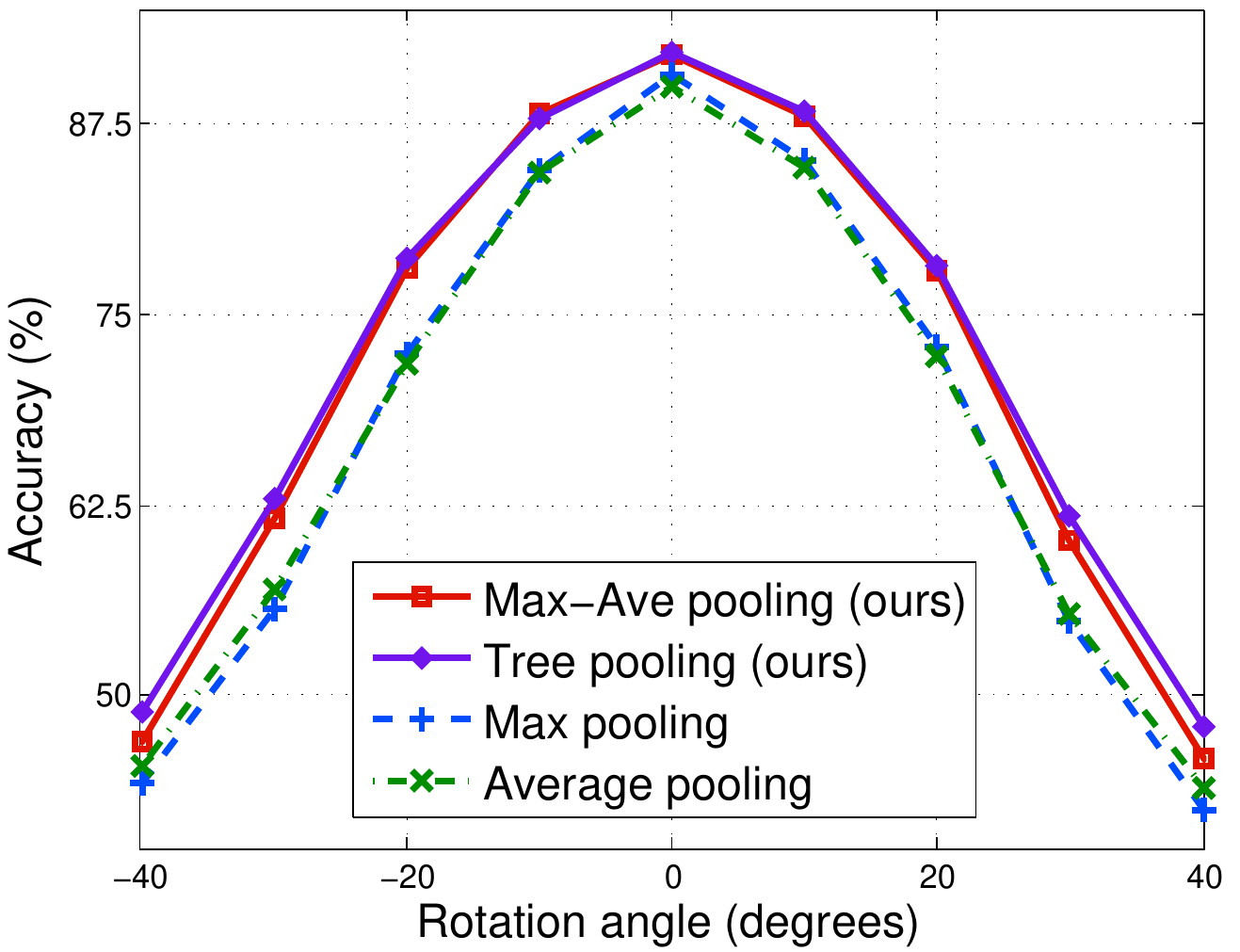}  
\includegraphics[width=0.33\linewidth]{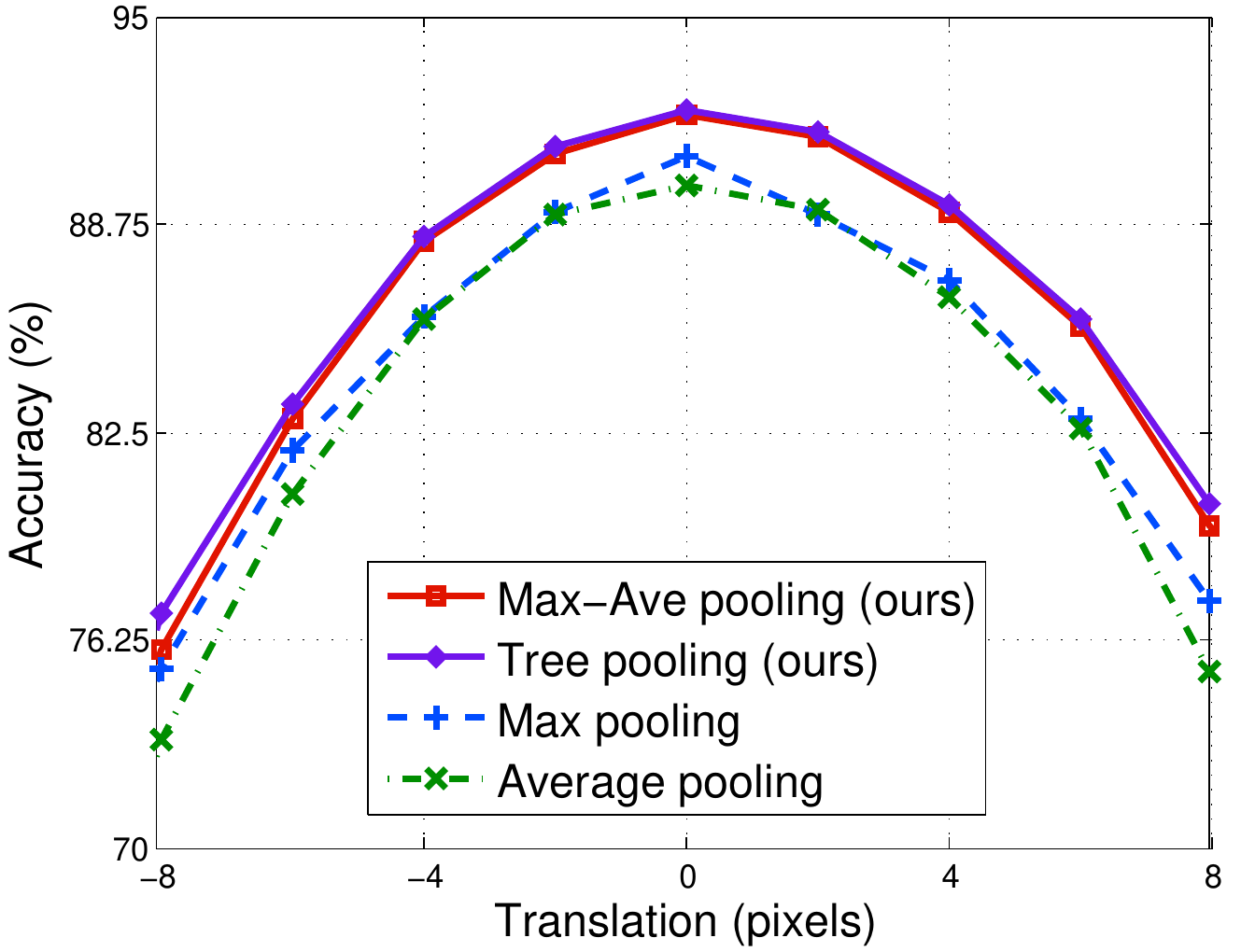}  
\includegraphics[width=0.33\linewidth]{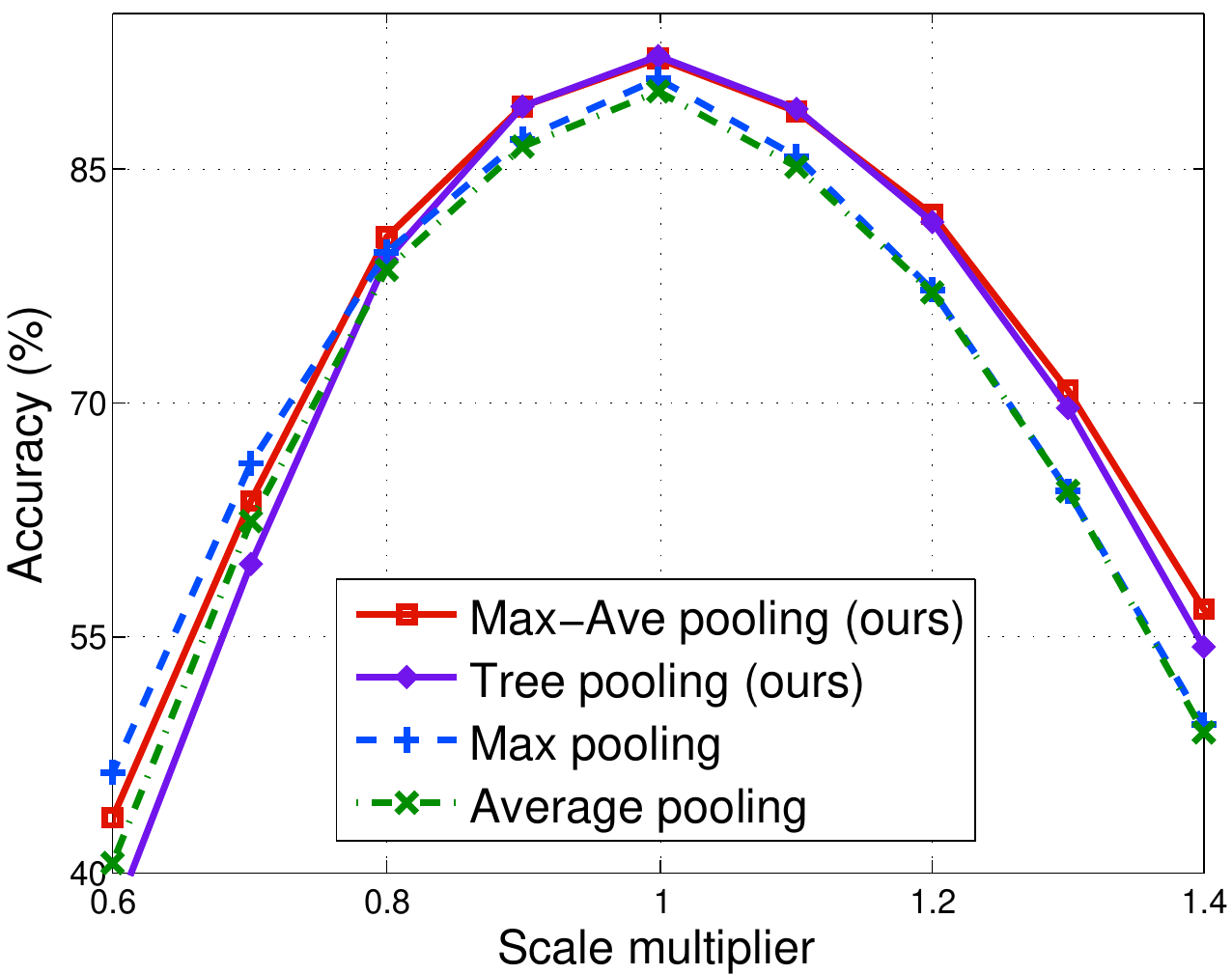}
\end{center}
\vspace{-5mm}
\caption{\label{fig:invariance} {\footnotesize{ Controlled experiment on CIFAR10 investigating 
the relative benefit of selected pooling operations in terms of robustness to three types of data variation.
The three kinds of variations we choose to investigate are rotation, translation,
and scale. With each kind of variation, we modify the CIFAR10 test images according to the listed amount.
We observe that, across all types and amounts of variation 
(except extreme down-scaling) the proposed pooling operations investigated here (gated max-avg and 2 level tree pooling) 
provide improved robustness to these transformations, relative to the standard 
choices of maxpool or avgpool.}}}
\vspace{-4mm}
\end{figure*}
\vspace{-4mm}
\subsubsection{Quick comparison: tree pooling}
\vspace{-2mm}
Table \ref{tab:tree} collects results related to tree pooling. We observe that 
on all datasets but the comparatively simple MNIST, adding a level to the tree pooling 
operation improves performance. However, even further benefit is obtained from the use 
of tree pooling in the first pooling layer and gated max-avg in the second. 

\begin{table}[!htp]
\vspace{-1mm}
\caption{\label{tab:tree} Classification error (in \%) comparison between our baseline model (trained
with conventional max pooling) and proposed methods involving tree pooling. A superscripted $^{+}$ 
indicates the standard data augmentation as in 
\cite{lin2013network,lee2015deeply,springenberg2015striving}.}
\begin{center}
\vspace{-4mm}
{\small
\begin{tabular}{c | c | c | c | c | c}
\hline
\shortstack{Method}                                                                  & {\tiny MNIST} & {\tiny CIFAR10} & {\tiny CIFAR10$^{+}$} & {\tiny CIFAR100} & {\tiny SVHN}  \\
\hline
Our baseline                                                                         & 0.39          & 9.10            & 7.32                  & 34.21            & 1.91          \\
\hline 
\shortstack{\textbf{\small Tree} \\ {\tiny 2 level; 1 per pool layer}}        & 0.35          & 8.25            & 6.88                  & 33.53            & 1.80          \\
\hline
\shortstack{\textbf{\small Tree} \\ {\tiny 3 level; 1 per pool layer}}        & 0.37          & 8.22            & 6.67                  & 33.13            & 1.70          \\
\hline
\shortstack{\textbf{\small Tree+Max-Avg} \\ {\tiny 1 per pool layer}}          & \textbf{0.31} & \textbf{7.62}   & \textbf{6.05}         & \textbf{32.37}   & \textbf{1.69} \\
\hline
\end{tabular}
}
\end{center}
\vspace{-3mm}
\end{table}

\textbf{Comparison with making the network deeper using conv layers}
To further investigate whether simply “adding depth” to our baseline network gives 
a performance boost comparable to that observed for our proposed pooling operations,
we report in Table \ref{tab:depth_comparison} below some additional experiments on CIFAR10 (error rate in percent; no 
data augmentation). If we count depth by counting any layer with learned parameters as an
extra layer of depth (even if there is only 1 parameter), the number of parameter 
layers in a baseline network with 2 additional standard convolution layers matches the number
of parameter layers in our best performing net (although the convolution layers contain many more parameters).

Our method requires only $72$ extra parameters and obtains state-of-the-art
$7.62\%$ error. On the other hand, making networks deeper with conv layers adds many more parameters but yields test error 
that does not drop below $9.08\%$ in the configuration explored. Since we follow each additional conv layer with a ReLU, 
these networks correspond to increasing nonlinearity as well as adding depth and adding (many) parameters. These experiments 
indicate that the performance of our proposed pooling is not accounted for as a simple effect of the addition of 
depth/parameters/nonlinearity.
\begin{table}[!htp]
\vspace{-4mm}
\footnotesize
\caption{\label{tab:depth_comparison} Classification error (\%) on CIFAR10 (without data augmentation) comparison between
networks made deeper with standard convolution layers and proposed Tree+(gated) Max-Avg pooling.}
\vspace{-4mm}
\begin{center}
\small
\begin{tabular}{l | c | c }
\hline
\shortstack{Method}                  & {\% Error}      & { \shortstack{Extra \\ parameters}}     \\
\hline
Baseline                             & 9.10          & 0           \\
w/ 1 extra conv layer (+ReLU)        & 9.08          & 0.6M        \\
w/ 2 extra conv layers (+ReLU)       & 9.17          & 1.2M        \\
w/ Tree+(gated) Max-Avg              & 7.62          & 72          \\
\hline
\end{tabular}
\end{center}
\vspace{-5mm}
\end{table}

\textbf{Comparison with alternative pooling layers}
To see whether we might find similar performance boosts by replacing the max pooling
in the baseline network configuration with alternative pooling operations such as 
stochastic pooling, ``pooling'' using a stride 2 convolution layer as pooling (cf 
All-CNN), or a simple fixed 50/50 proportion in max-avg pooling, we performed another 
set of experiments on unaugmented CIFAR10. From the baseline error rate of 9.10\%, 
replacing each of the 2 max pooling layers with stacked stride 2 conv:ReLU (as in \cite{springenberg2015striving})
lowers the error to 8.77\%, but adds 0.5M extra parameters. Using stochastic pooling \cite{zeiler2013stochastic} 
adds computational overhead but no parameters and results in 8.50\% error. A simple
50/50 mix of max and average is computationally light and yields 8.07\% error with no
additional parameters. Finally, our tree+gated max-avg configuration adds 72 parameters
and achieves a state-of-the-art 7.62\% error.
\vspace{-2mm}
\section{Quick Performance Overview}
\vspace{-4mm}
For ease of discussion, we collect here observations from subsequent experiments 
with a view to highlighting aspects that shed light on the performance characteristics
of our proposed pooling functions. 

\vspace{-.5mm}
First, as seen in the experiment shown in Figure \ref{fig:invariance} replacing 
standard pooling operations with either gated max-avg or (2 level) tree pooling 
(each using the ``one per layer'' option) yielded a boost (relative to max or avg pooling) 
in CIFAR10 test accuracy as the test images underwent three different kinds of transformations. 
This boost was observed across the entire range of transformation amounts for each 
of the transformations (with the exception of extreme downscaling). We already observe improved 
robustness in this initial experiment and intend to investigate more instances 
of our proposed pooling operations as time permits.

\vspace{-.5mm}
Second, the performance that we attain in the experiments reported in
Figure \ref{fig:invariance}, Table \ref{tab:mixed_and_gated}, Table \ref{tab:tree}, Table \ref{tab:all}, and
Table \ref{tab:imagenet} is achieved with very modest additional
numbers of parameters --- e.g. on CIFAR10, our best performance (obtained with the 
tree+gated max-avg configuration) only uses an additional $72$ parameters 
(above the ~1.8M of our baseline network) and yet reduces test error from $9.10\%$ 
to $7.62\%$; see the \textbf{CIFAR10} Section for details. In our AlexNet experiment, replacing the 
maxpool layers with our proposed pooling operations gave a $6\%$ relative reduction 
in test error (top-5, single-view) with only 45 additional parameters (above the $>$50M 
of standard AlexNet); see the \textbf{ImageNet 2012} Section for details. We also 
investigate the additional time incurred when using our proposed pooling operations;
in the experiments reported in the \textbf{Timing} section, this overhead ranges from $5\%$ to $15\%$.

\vspace{-.7mm}
\textbf{Testing invariance properties}
Before going to the overall classification results, we investigate the invariance
properties of networks utilizing either standard pooling operations (max and average)
or two instances of our proposed pooling operations (gated max-avg and 2 level tree, 
each using the ``1 per pool layer'' option) that we find to yield best performance 
(see Sec. \ref{sec:exp} for architecture details used across each network). We begin 
by training four different networks on the CIFAR10 training set, one for each of 
the four pooling operations selected for consideration; training details are found in Sec. 
\ref{sec:exp}. We seek to determine the respective invariance properties of these 
networks by evaluating their accuracy on various transformed versions of 
the CIFAR10 test set. Figure \ref{fig:invariance} illustrates the test accuracy 
attained in the presence of image rotation, (vertical) translation, and scaling 
of the CIFAR10 test set.  

\textbf{Timing}
In order to evaluate how much additional time is incurred by the use of our proposed learned pooling operations, we 
measured the average forward+backward time per CIFAR10 image. In each case, the one per layer option is used. We find 
that the additional computation time incurred ranges from $5\%$ to $15\%$. More specifically, the baseline
network took 3.90 ms; baseline with mixed max-avg took 4.10 ms; baseline with gated max-avg took 4.16 ms; baseline with 
2 level tree pooling took 4.25 ms; finally, baseline with tree+gated max-avg took 4.46 ms. 
\vspace{-4mm}
\section{Experiments} \label{sec:exp}
\vspace{-4mm}
We evaluate the proposed max-average pooling and tree pooling approaches on five 
standard benchmark datasets: MNIST \cite{lecun1998mnist}, CIFAR10 \cite{krizhevsky2009learning}, 
CIFAR100 \cite{krizhevsky2009learning}, SVHN \cite{netzer2011reading} and ImageNet \cite{russakovsky2014imagenet}. 
To control for the effect of differences in data or data preparation, we match our data and data preparation to that 
used in \cite{lee2015deeply}. Please refer to \cite{lee2015deeply} for the detailed description.
 
We now describe the basic network architecture and then will specify the various hyperparameter 
choices. The basic experiment architecture contains six  $3\times3$  standard convolutional layers (named conv1 to conv6) and three 
mlpconv layers (named mlpconv1 to mlpconv3) \cite{lin2013network}, placed after 
conv2, conv4, and conv6, respectively. We chose the number of channels at each layer
to be analogous to the choices in \cite{lin2013network,lee2015deeply}; the specific numbers 
are provided in the sections for each dataset. We follow every one of these conv-type 
layers with ReLU activation functions. One final mlpconv layer  (mlpconv4) is used 
to reduce the dimension of the last layer to match the total number of classes for 
each different dataset, as in \cite{lin2013network}. The overall model has parameter 
count analogous to \cite{lin2013network,lee2015deeply}. The proposed max-average pooling 
and tree pooling layers with $3\times3$ pooling 
regions are used after mlpconv1 and mlpconv2 layers \footnote{There is one exception: 
on the very small images of the MNIST dataset, the second pooling layer uses $2\times2$ pooling regions.}. 
We provide a detailed listing of the network configurations in Table \ref{tab:network_configurations} 
in the Supplementary Materials. 

Moving on to the hyperparameter settings, dropout with rate $0.5$ is used after each 
pooling layer. We also use hidden layer supervision to ease the training process as 
in \cite{lee2015deeply}. The learning rate is decreased whenever the validation error 
stops decreasing; we use the schedule $\{0.025,0.0125,0.0001\}$ for all experiments. 
The momentum of $0.9$ and weight decay of $0.0005$ are fixed for all datasets as 
another regularizer besides dropout. All the initial pooling filters and pooling masks 
have values sampled from a Gaussian distribution with zero mean and standard deviation 
$0.5$. We use these hyperparameter settings for all experiments reported in Tables \ref{tab:mixed_and_gated}, 
\ref{tab:tree}, and \ref{tab:depth_comparison}. 
No model averaging is done at test time. 

\vspace{-4mm}
\subsection{Classification results}
\vspace{-3mm}
Tables \ref{tab:mixed_and_gated} and \ref{tab:tree} show our overall experimental results. Our baseline is a
network trained with conventional max pooling. \emph{Mixed} refers 
to the same network but with a max-avg pooling strategy in both the first and second 
pooling layers (both using the mixed strategy); \emph{Gated} has a corresponding meaning.
\emph{Tree} (with specific number of levels noted below) refers to the same again, but with our tree 
pooling in the first pooling layer only; we do not see further improvement when tree pooling is 
used for both pooling layers.  This observation motivated us to consider following a tree pooling 
layer with a gated max-avg pooling layer: \emph{Tree+Max-Average} refers to a network 
configuration with (2 level) tree pooling for the first pooling layer and gated max-average pooling 
for the second pooling layer. 
All results are produced from the same network structure 
and hyperparameter settings --- the only difference is in the choice of pooling function. 
See Table \ref{tab:network_configurations} for details.

\textbf{MNIST}
Our MNIST model has $\{128,\allowbreak 128,\allowbreak 192,\allowbreak 192,\allowbreak 256,\allowbreak 256\}$ channels 
for conv1 to conv6 and $\{128,\allowbreak 192,\allowbreak 256\}$ channels for mlpconv1 to mlpconv3, respectively. 
Our only preprocessing is mean subtraction. Tables \ref{tab:all},\ref{tab:mixed_and_gated}, and \ref{tab:tree} show previous 
best results and those for our proposed pooling methods.

\textbf{CIFAR10}
Our CIFAR10 model has $\{128,\allowbreak 128,\allowbreak 192,\allowbreak 192,\allowbreak 256,\allowbreak 256\}$ channels for conv1 to conv6 
and $\{128,\allowbreak 192,\allowbreak 256\}$ channels for mlpconv1 to mlpconv3, respectively. 
We also performed an experiment in which we learned a single pooling filter
without the tree structure (i.e., a singleton leaf node containing $9$ parameters; 
one such singleton leaf node per pooling layer) and obtained $0.3\%$ improvement over 
the baseline model. Our results indicate that performance improves when the pooling filter 
is learned, and further improves when we also learn how to combine learned pooling filters.

The All-CNN method in \cite{springenberg2015striving} uses convolutional layers 
in place of pooling layers in a CNN-type network architecture. However, a standard 
convolutional layer requires many more parameters than a gated max-average pooling 
layer (only $9$ parameters for a $3 \times 3$ pooling region kernel size in the 1 
per pooling layer option) or a tree-pooling layer ($27$ parameters for a 2 level 
tree and $3 \times 3$ pooling region kernel size, again in the 1 per pooling layer 
option).  The pooling operations in our tree+max-avg network configuration use $7\times9 = 63$ 
parameters for the (first, 3 level) tree-pooling layer --- 4 leaf nodes and 3 internal 
nodes --- and $9$ parameters in the gating mask used for the (second) gated max-average pooling 
layer, while the best result in \cite{springenberg2015striving} contains a total 
of nearly $500,000$ parameters in layers performing ``pooling like'' operations; 
the relative CIFAR10 accuracies are $7.62\%$ (ours) and $9.08\%$ (All-CNN). 

For the data augmentation experiment, we followed the standard data augmentation procedure 
\cite{lin2013network,lee2015deeply,springenberg2015striving}.
When training with augmented data, we observe the same trends seen in the ``no data augmentation'' experiments.
We note that \cite{graham2014fractional} reports a $4.5\%$ error rate with extensive
data augmentation (including translations, rotations, reflections, stretching, and
shearing operations) in a much wider and deeper $50$ million parameter network
--- $28$ times more than are in our networks.

\begin{table}[t]
\footnotesize
\caption{\label{tab:all} Classification error (in \%) reported by recent comparable publications
on four benchmark datasets with a single model and no data augmentation, unless 
otherwise indicated. A superscripted $^{+}$ indicates the standard data augmentation
as in \cite{lin2013network,lee2015deeply,springenberg2015striving}.
A ``-'' indicates that the cited work did not report results for that dataset. 
A fixed network configuration using the proposed tree+max-avg pooling
(1 per pool layer option) yields state-of-the-art performance on all datasets (with the exception of CIFAR100).
}
\vspace{-3mm}
\begin{center}
\begin{tabular}{l| c | c | c | c | c}
\hline
Method                                                      & {\tiny MNIST} & {\tiny CIFAR10} & {\tiny CIFAR10$^{+}$}  & {\tiny CIFAR100} & {\tiny SVHN}          \\
\hline
{\tiny CNN   \cite{jarrett2009best}}                         & 0.53          & -               & -                      &  -              & -             \\
{\tiny Stoch. Pooling \cite{zeiler2013stochastic}}          & 0.47          & 15.13           & -                      & 42.51           & 2.80          \\
{\tiny Maxout Networks \cite{goodfellow2013maxout}}         & 0.45          & 11.68           & 9.38                   & 38.57           & 2.47          \\
{\tiny Prob. Maxout \cite{springenberg2014improving}}       & -             & 11.35           & 9.39                   & 38.14           & 2.39          \\
{\tiny Tree Priors \cite{srivastava2013discriminative}}     & -             & -               & -                      & 36.85           & -             \\
{\tiny DropConnect \cite{li2013regularization}}             & 0.57          &  9.41           & 9.32                   & -               & 1.94          \\
{\tiny FitNet \cite{romero2015fitnets}}                     & 0.51          & -               & 8.39                   & 35.04           & 2.42          \\
{\tiny NiN \cite{lin2013network}}                           & 0.47          & 10.41           & 8.81                   & 35.68           & 2.35          \\
{\tiny DSN \cite{lee2015deeply}}                            & 0.39          & 9.69            & 7.97                   & 34.57           & 1.92          \\
{\tiny NiN + LA units \cite{agostinelli2015learning}}       & -             & 9.59            & 7.51                   & 34.40           & -             \\
{\tiny dasNet \cite{stollenga2014deep}}                     & -             & 9.22            & -                      & 33.78           & -             \\
{\tiny All-CNN \cite{springenberg2015striving}}             & -             & 9.08            & 7.25                   & 33.71           & -             \\
{\tiny R-CNN \cite{liang2015recurrent}}                     & 0.31          & 8.69            & 7.09                   & \textbf{31.75}  & 1.77          \\
\hline
{\tiny Our baseline}                                        & 0.39          & 9.10            & 7.32                   & 34.21           & 1.91          \\
\hline
\shortstack{{\tiny Our Tree+Max-Avg}} & \textbf{0.31} & \textbf{7.62}   & \textbf{6.05}          & 32.37           & \textbf{1.69} \\
\hline
\end{tabular}
\end{center}
\vspace{-5mm}
\end{table}

\vspace{-.5mm}
\textbf{CIFAR100} 
Our CIFAR100 model has $192$ channels for all convolutional layers and $\{96,192,192\}$
channels for mlpconv1 to mlpconv3, respectively. 

\vspace{-.5mm}
\textbf{Street view house numbers}
Our SVHN model has $\{128,\allowbreak 128,\allowbreak  320,\allowbreak  320,\allowbreak  384,\allowbreak  384\}$ channels for conv1 to conv6 and
$\{96,\allowbreak 256,\allowbreak 256\}$ channels for mlpconv1 to mlpconv3, respectively. 
In terms of amount of data, SVHN has a larger training data set
($>$600k versus the $\approx$50k of most of the other benchmark  datasets). The much 
larger amount of training data motivated us to explore what performance we might observe if 
we pursued the one per layer/channel/region option, which even for the simple mixed max-avg strategy 
results in a huge increase in total the number of parameters to learn in our proposed pooling layers: 
specifically, from a total of 2 in the mixed max-avg strategy, 1 parameter per pooling 
layer option, we increase to 40,960. 

\vspace{-.5mm}
Using this one per layer/channel/region option for the mixed 
max-avg strategy, we observe test error (in $\%$) of 0.30 on MNIST, 8.02 on CIFAR10, 
6.61 on CIFAR10$^+$, 33.27 on CIFAR100, and 1.64 on SVHN. Interestingly, 
for MNIST, CIFAR10$^+$, and CIFAR100 this mixed max-avg (1 per layer/channel/region)
performance is between mixed max-avg (1 per layer) and gated max-avg (1 per layer); 
on CIFAR10 mixed max-avg (1 per layer/channel/region) is worse than either of the 
1 per layer max-avg strategies.  The SVHN result using mixed max-avg 
(1 per layer/channel/region) sets a new state of the art. 

\vspace{-.5mm}
\textbf{ImageNet 2012}
In this experiment we do not directly compete with the best performing result in
the challenge (since the winning methods \cite{szegedy2014going} involve many 
additional aspects beyond pooling operations), but rather to provide an illustrative comparison of the relative 
benefit of the proposed pooling methods versus conventional max pooling on this
dataset. We use the same network structure and parameter setup as in \cite{krizhevsky2012imagenet}
(no hidden layer supervision)  but simply replace the first max pooling with the (proposed 2 level) tree pooling 
(2 leaf nodes and 1 internal node for $27=3\times9$ parameters) and replace the second and third max pooling with gated
max-average pooling (2 gating masks for $18=2\times9$ parameters). Relative to the original AlexNet,
this adds $45$ more parameters (over the $>$50M in the original) and achieves relative 
error reduction of $6\%$ (for top-5, single-view) and  $5\%$ (for top-5, multi-view). Our GoogLeNet 
configuration uses 4 gated max-avg pooling layers, for a total of 36 extra parameters over 
the 6.8 million in standard GoogLeNet.
Table \ref{tab:imagenet} shows a direct comparison (in each case we use single 
net predictions rather than ensemble).  

\begin{table}[!htp]
\vspace{-3mm}
{\footnotesize
\begin{center}
\caption{\label{tab:imagenet} ImageNet 2012 test error (in \%). BN denotes Batch Normalization \cite{ioffe2015batch}.}
\vspace{-3mm}
\begin{tabular}{lcccc} 
\multicolumn{1}{l}{Method}                  &\shortstack{top-1 \\ s-view} & \shortstack{top-5 \\ s-view} &\shortstack{top-1 \\ m-view} & \shortstack{top-5 \\ m-view}
\\ \hline 
\scriptsize{AlexNet \cite{krizhevsky2012imagenet} }      & 43.1  & 19.9  & 40.7   & 18.2 \\
\scriptsize{AlexNet w/ ours  }                           & 41.4  & 18.7  & 39.3   & 17.3 \\
\hline 
\scriptsize{GoogLeNet \cite{szegedy2014going}  }         & -     & 10.07 & -      & 9.15 \\
\scriptsize{GoogLeNet w/ BN    }                         & 28.68 &  9.53 & 27.81  & 9.09 \\
\scriptsize{GoogLeNet w/ BN + ours }                     & 28.02 &  9.16 & 27.60  & 8.93 \\ 
\hline 
\end{tabular}
\end{center}
}
\vspace{-4mm}
\end{table}

\vspace{-4mm}
\section{Observations from Experiments}
\vspace{-4mm}
In each experiment, using any of our proposed pooling operations boosted performance.
A fixed network configuration using the proposed tree+max-avg pooling (1
per pool layer option) yields state-of-the-art performance
on MNIST, CIFAR10 (with and without data augmentation), and SVHN.
We observed boosts in tandem with data augmentation, multi-view predictions, batch normalization,
and several different architectures --- NiN-style, DSN-style, the $>$50M parameter AlexNet, and the 22-layer GoogLeNet.

\noindent {\bf Acknowledgment} This work is supported by NSF awards IIS-1216528 (IIS-1360566) and IIS-0844566(IIS-1360568).

{
\footnotesize
\bibliographystyle{ieee}
\bibliography{merged}
}

\clearpage
\setcounter{section}{0}
\renewcommand{\thesection}{A\arabic{section}}
\setcounter{table}{0}
\renewcommand{\thetable}{A\arabic{table}}
\setcounter{figure}{0}
\renewcommand{\thefigure}{A\arabic{figure}}

\section{Supplementary Materials}

\begin{figure*}[!htp]
\begin{center}
\includegraphics[width=0.015\linewidth]{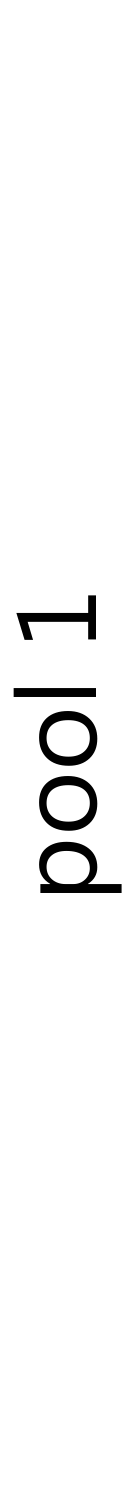} 
\includegraphics[width=0.235\linewidth]{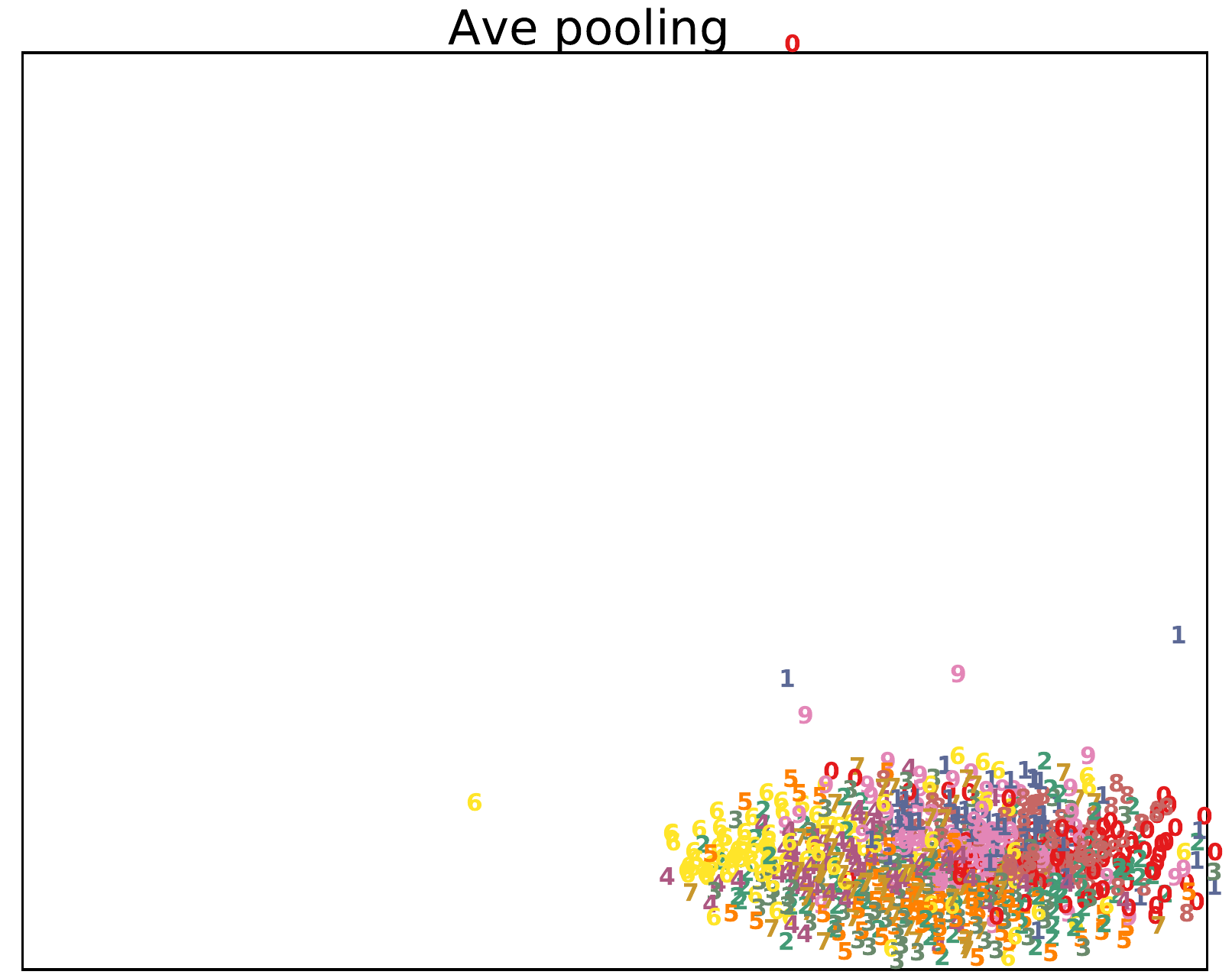}  
\includegraphics[width=0.235\linewidth]{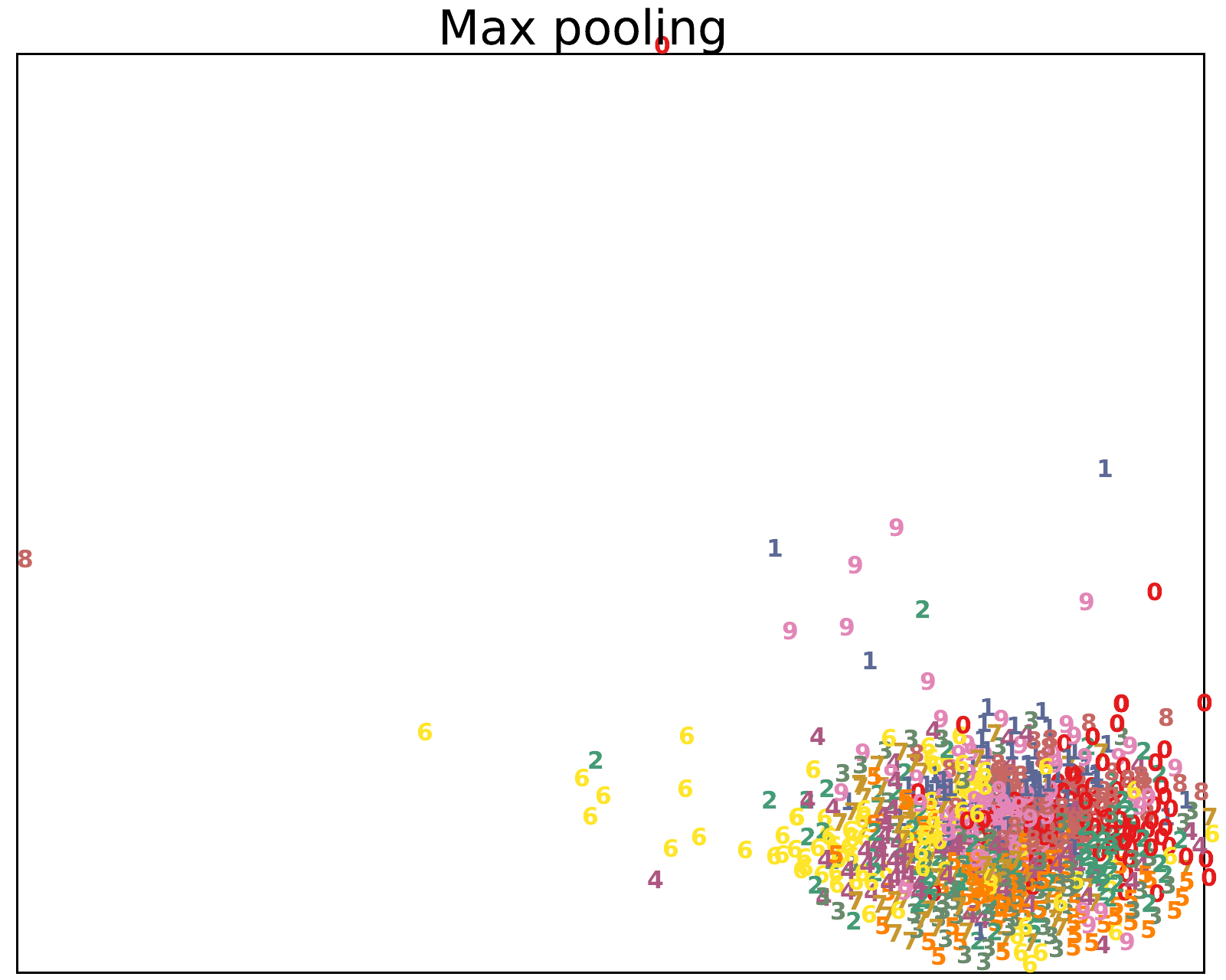} 
\includegraphics[width=0.235\linewidth]{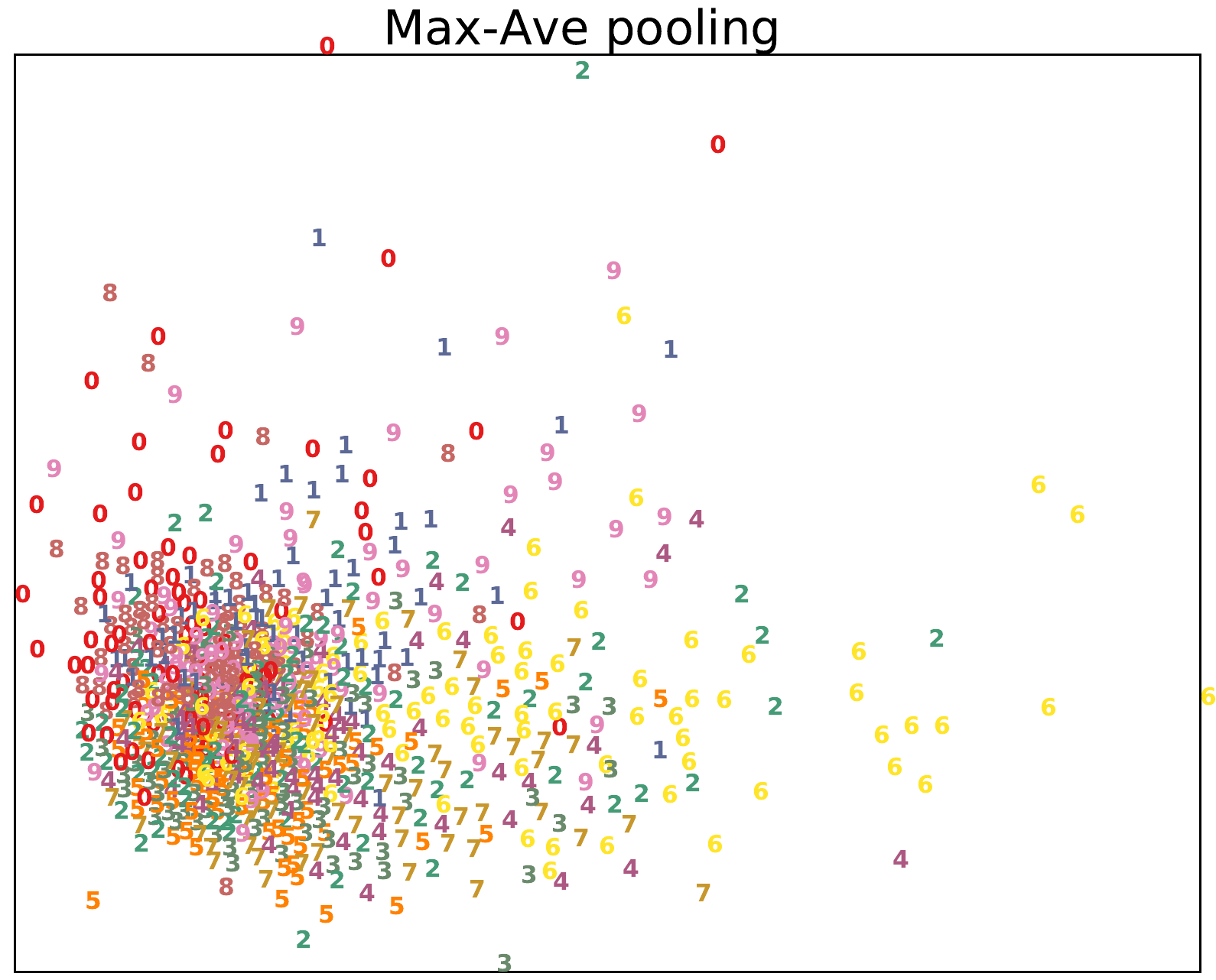} 
\includegraphics[width=0.235\linewidth]{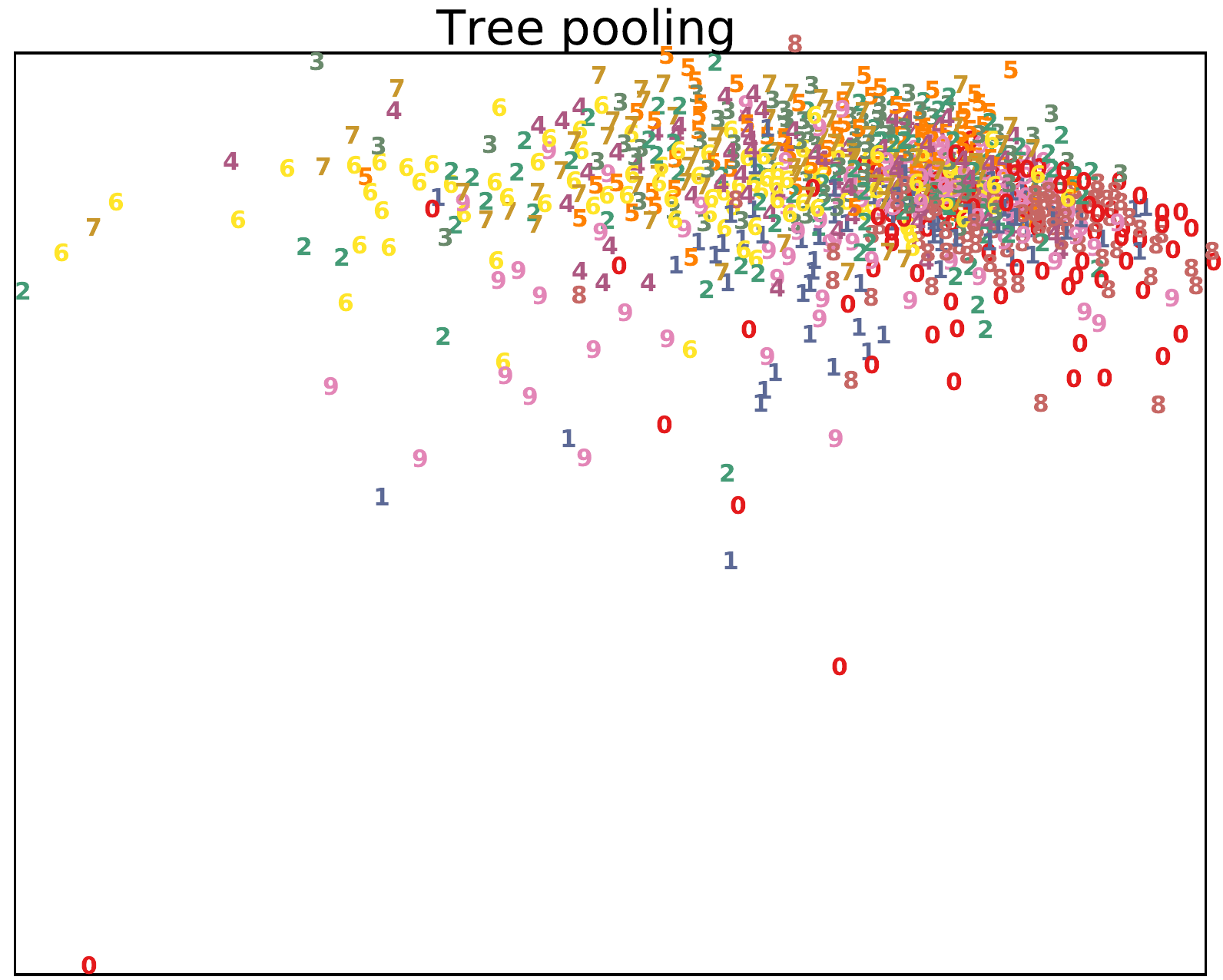}  \\
\includegraphics[width=0.015\linewidth]{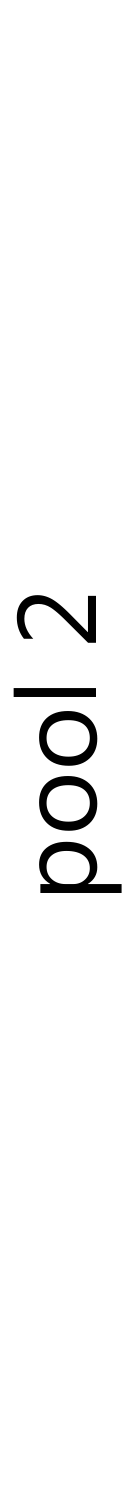}
\includegraphics[width=0.235\linewidth]{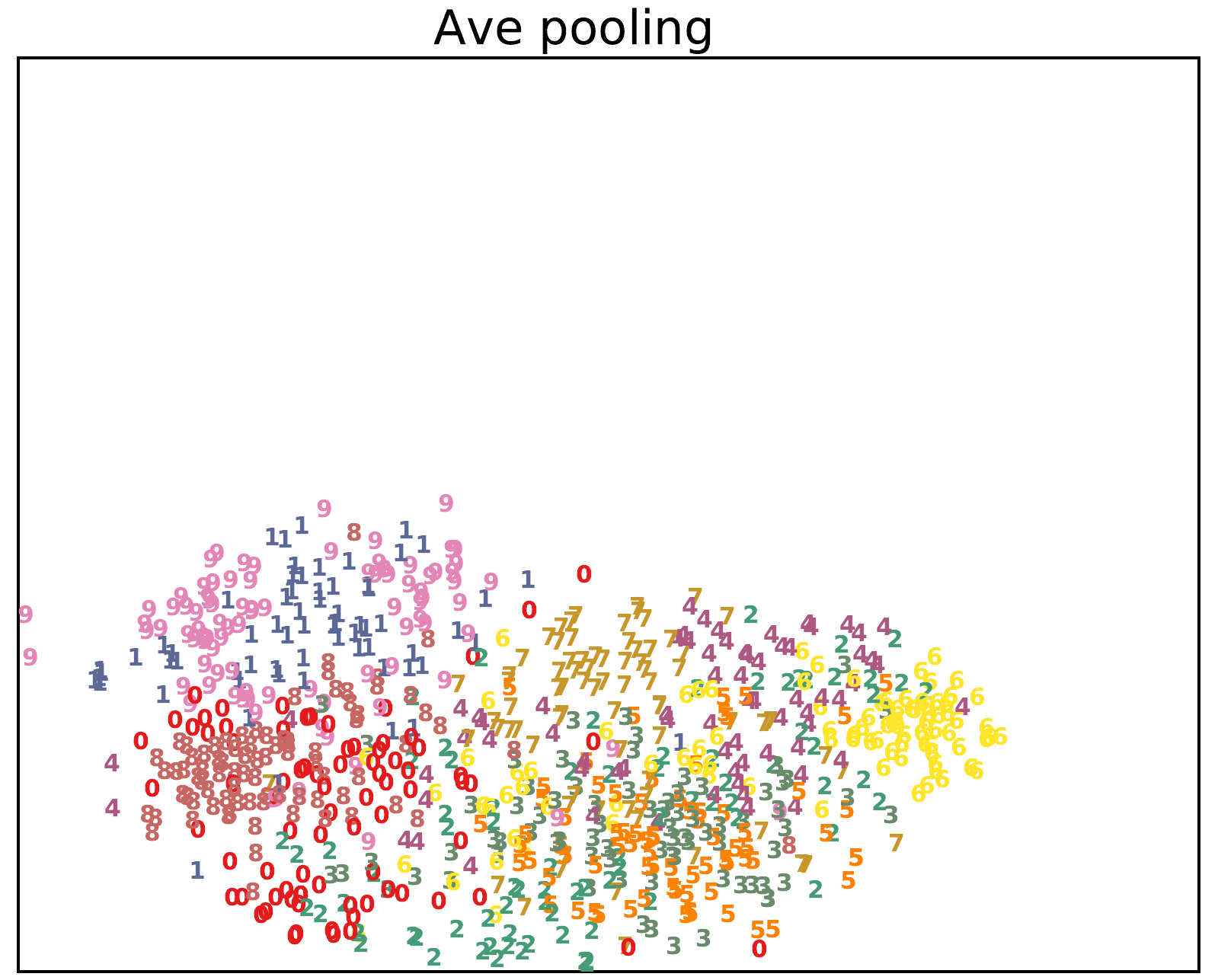}  
\includegraphics[width=0.235\linewidth]{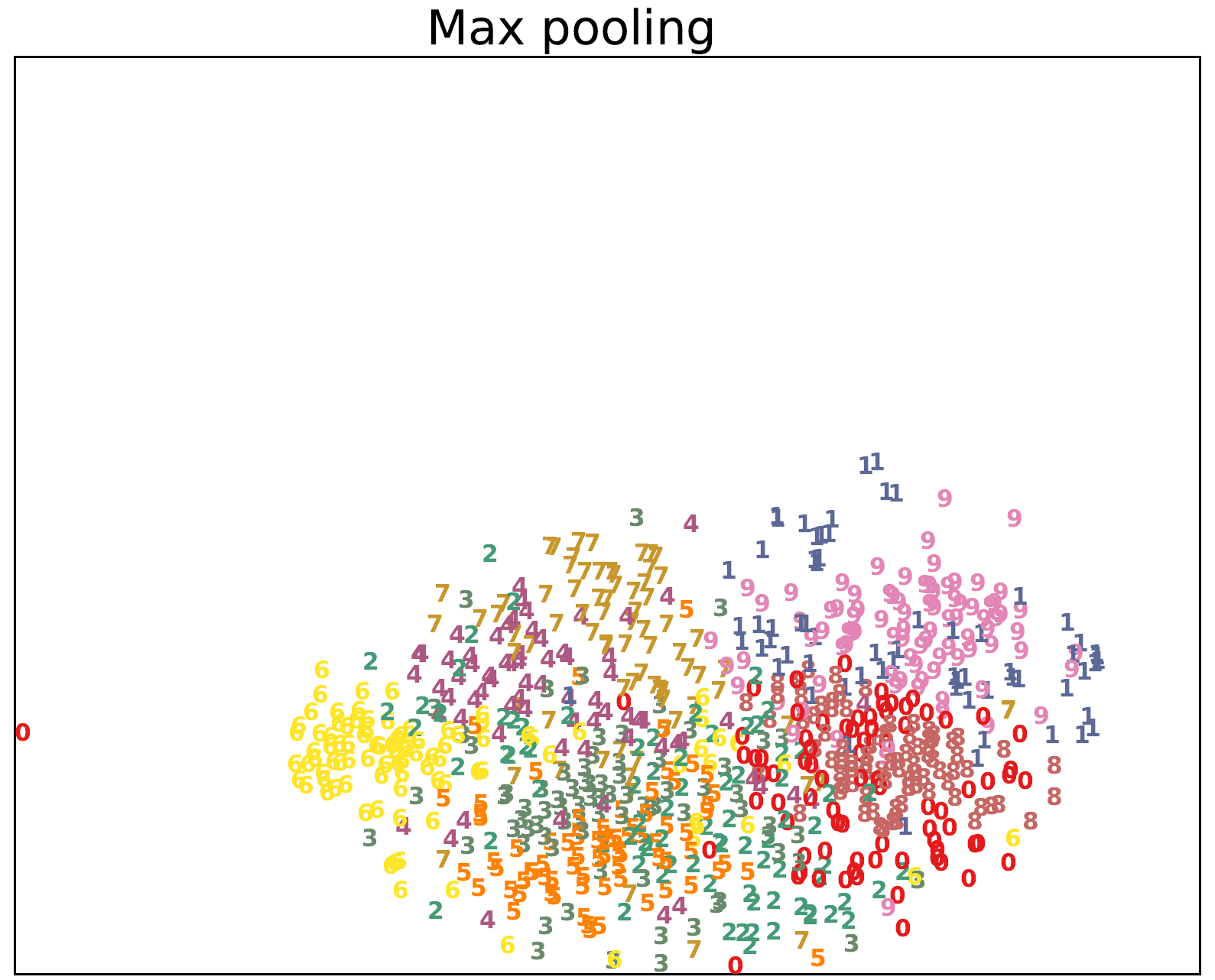} 
\includegraphics[width=0.235\linewidth]{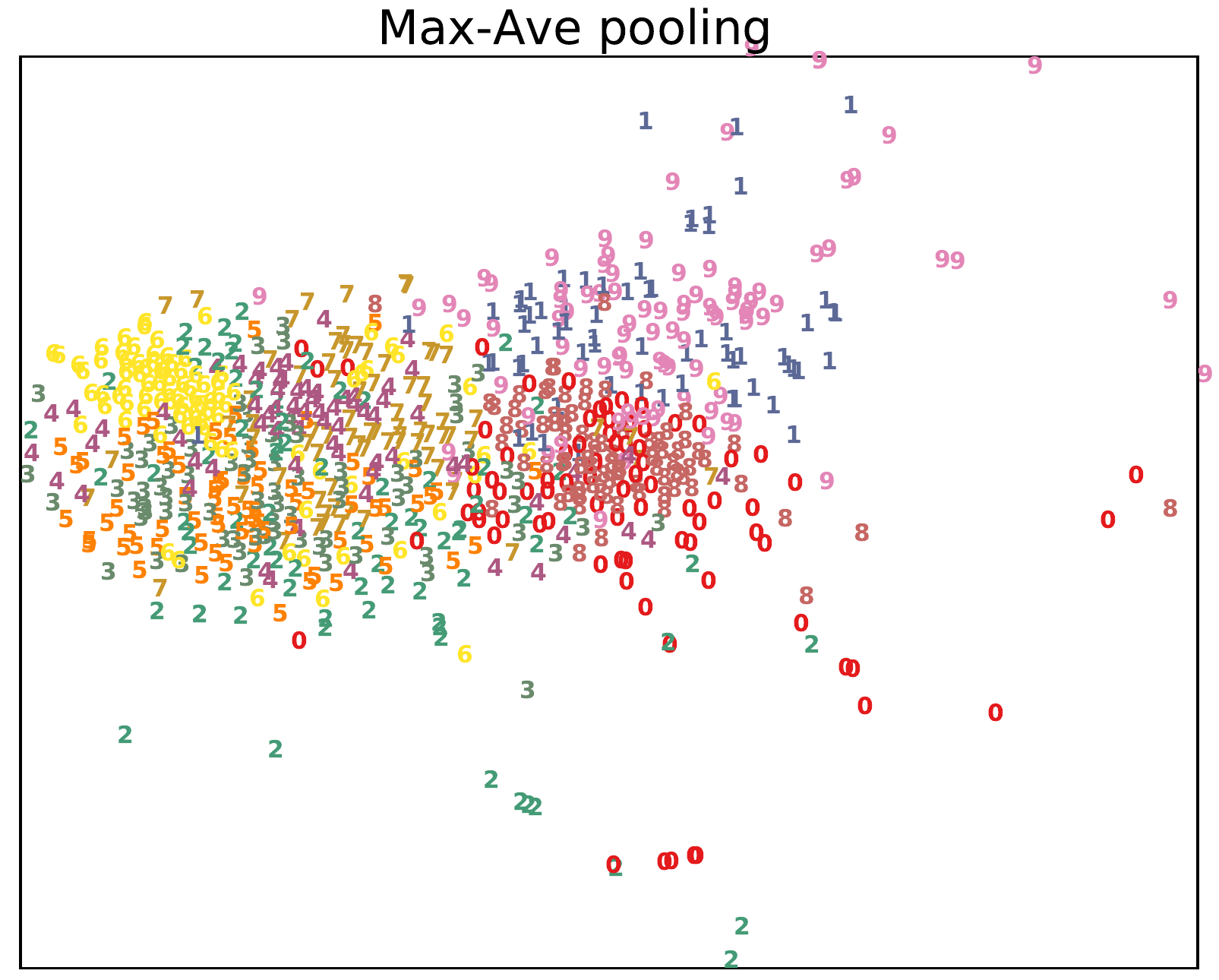} 
\includegraphics[width=0.235\linewidth]{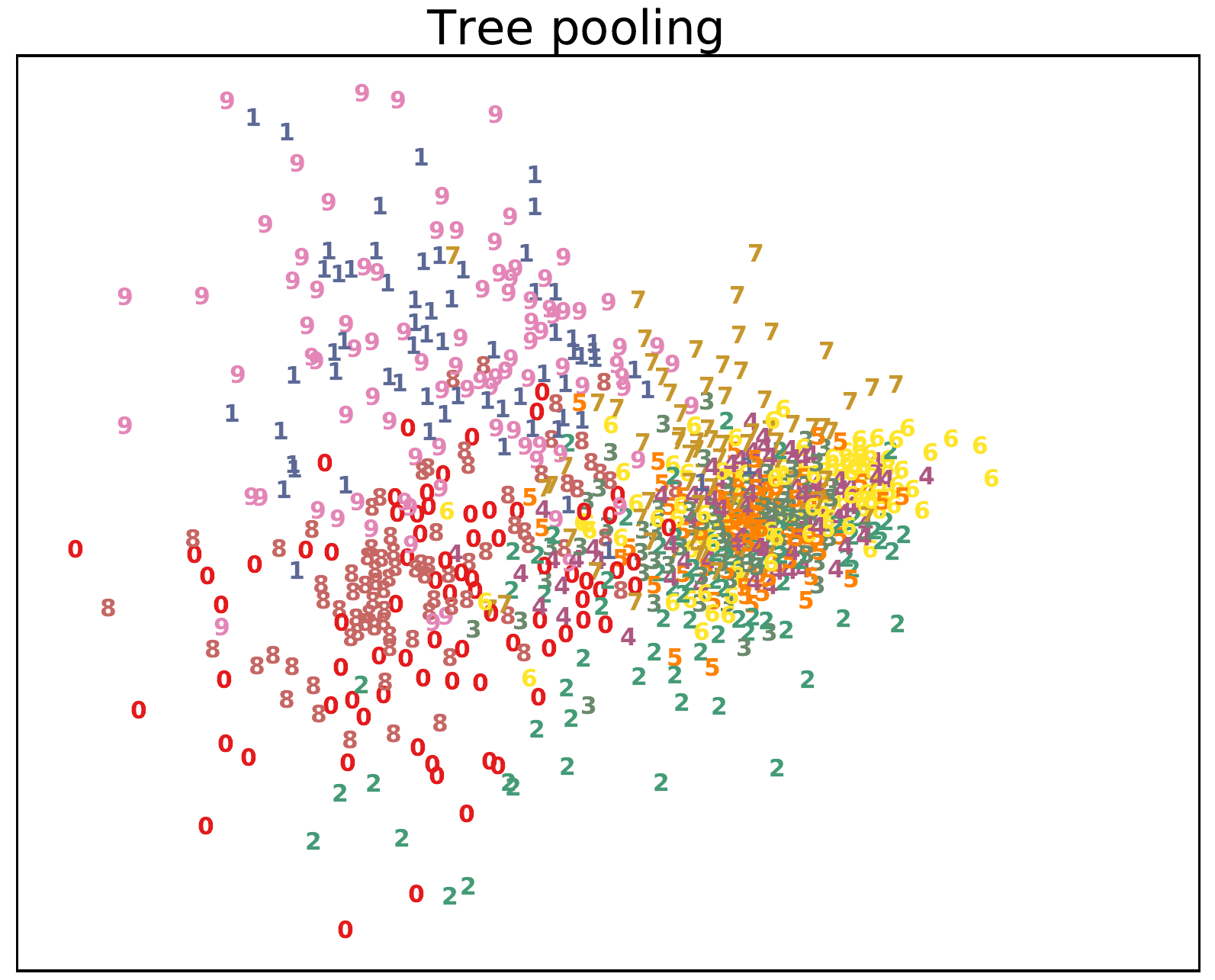}
\end{center}
\caption{t-SNE embeddings of the output responses from different pooling operations on
the CIFAR10 test set (with classes indicated). From left to right: average, max,
gated max-avg, and (2 level) tree pooling. The first and the second rows show the
first and the second pooling layers, respectively. Best viewed in color.}
\label{fig-embedding}
\end{figure*}

\textbf{Visualization of network internal representations}
To gain additional qualitative understanding of the pooling methods we are considering,
we use the popular t-SNE \cite{vandermaaten2008visualizing} algorithm to visualize 
embeddings of some internal feature responses from pooling operations. Specifically,
we again use four networks (one utilizing each of the selected types of pooling) trained on the CIFAR10
training set (see Sec. \ref{sec:exp} for architecture details used across each network).
We extract feature responses for a randomly chosen $800$-image subset of the CIFAR10 
test set at the first (i.e., earliest) and second pooling layers of each network.  These 
feature response vectors are then embedded into 2-d using t-SNE; see Figure \ref{fig-embedding}. 

The first row shows the embeddings of the internal activations immediately after the first 
pooling operation; the second row shows embeddings of activations immediately after the 
second pooling operation. From left to right we plot the t-SNE embeddings of the pooling 
activations within networks that are trained with average, max, gated max-avg, and (2 level) 
tree pooling. We can see that certain classes such as ``0'' (airplane), ``2'' (bird), 
and ``9'' (truck) are more separated with the proposed methods than they are with 
the conventional average and max pooling functions. We can also see that the embeddings 
of the second-pooling-layer activations are generally more separable than the 
embeddings of first-pooling-layer activations.

\begin{table*}[ht]
\vspace{-20mm}
\begin{centering}
{ \caption{\label{tab:network_configurations} Here we provide explicit statement of the experimental
conditions (specifically, network layer configurations) explored in
Tables \ref{tab:mixed_and_gated}, \ref{tab:tree}, and 
\ref{tab:all}. We list all conv-like layers and pool-like layers,
but ReLUs are suppressed to lighten the amount of text; these follow
each standard conv layer. Also, all network configurations incorporate
deep supervision after each standard convolution layer; this is also
suppressed for clarity.  We bold the changes made to the baseline
DSN layer configuration. We now describe the meaning of
entries in the table. Each column in the table lists the sequence
of layer types used in that network configuration. When a row cell
spans multiple columns (i.e. configurations), this indicates that
the layer type listed in that cell is kept the same across the corresponding
network configurations. Thus, every network in our experiments begins
with a stacked pair of 3x3 (standard) conv layers followed by a 1x1
mlpconv layer. For a specific example, let us consider the network
configuration in the column headed ``mixed max-avg'' - the sequence
of layers in this configuration is: 3x3 (standard) conv, 3x3 (standard)
conv, 1x1 mlpconv, \textbf{3x3 mixed max-avg pool},
3x3 (standard) conv, 3x3 (standard) conv, 1x1 mlpconv, \textbf{3x3
mixed max-avg pool}, 3x3 (standard) conv, 3x3 (standard)
conv, 1x1 mlpconv, 1x1 mlpconv, 8x8 global vote (cf. \cite{lin2013network}) (we again 
omit mention of ReLUs and deep supervision). CIFAR100 uses (2 level) tree+max-avg; CIFAR10 uses (3 level) 
tree+max-avg. As a final note: for the MNIST experiments only, the second pooling operation 
uses 2x2 regions instead of the 3x3 regions used on the other datasets.}
}
{\scriptsize
\begin{tabular}{c|c|c|c|c|c}
\hline 
\multicolumn{6}{c}{Network layer configurations reported in Tables \ref{tab:mixed_and_gated}, \ref{tab:tree}, and 
\ref{tab:all} of the main paper.}\tabularnewline
\hline 
DSN (baseline) & mixed max-avg & gated max-avg & 2 level tree pool & 3 level tree pool & tree+gated max-avg pool \tabularnewline 
\hline 
\hline 
\multicolumn{6}{c}{3x3 (standard) conv }\tabularnewline
\hline 
\multicolumn{6}{c}{3x3 (standard) conv }\tabularnewline
\hline 
\multicolumn{6}{c}{1x1 mlpconv}\tabularnewline
\hline 
\multirow{1}{*}{3x3 maxpool} & \textbf{{3x3 mixed max-avg}} & \textbf{{3x3 gated max-avg}} & \textbf{{3x3 2 level tree pool}} & \textbf{{3x3 3 level tree pool}} & \textbf{{3x3 2/3 level tree pool}}\tabularnewline
\hline 
\multicolumn{6}{c}{{{3x3 (standard) conv }}}\tabularnewline
\hline 
\multicolumn{6}{c}{{{3x3 (standard) conv }}}\tabularnewline
\hline 
\multicolumn{6}{c}{{{1x1 mlpconv}}}\tabularnewline
\hline 
{{3x3 maxpool}} & \textbf{{3x3 mixed max-avg}} & \textbf{{3x3 gated max-avg}} & 3x3 maxpool & 3x3 maxpool & \textbf{{3x3 gated max-avg}}\tabularnewline
\hline 
\multicolumn{6}{c}{{{3x3 (standard) conv }}}\tabularnewline
\hline 
\multicolumn{6}{c}{{{3x3 (standard) conv }}}\tabularnewline
\hline 
\multicolumn{6}{c}{{{1x1 mlpconv}}}\tabularnewline
\hline 
\multicolumn{6}{c}{{{1x1 mlpconv}}}\tabularnewline
\hline 
\multicolumn{6}{c}{{{8x8 global vote}}}\tabularnewline
\end{tabular}
}
\par
\end{centering}
\end{table*}
 
\end{document}